\documentclass[10pt,twocolumn,letterpaper,pagebackref,breaklinks,colorlinks,allcolors=cvprblue]{article}

\usepackage[pagenumbers]{cvpr}              
\definecolor{cvprblue}{rgb}{0.21,0.49,0.74}

\usepackage{hyperref}
\usepackage{adjustbox}
\usepackage{tabularx}
\usepackage{makecell}
\usepackage{subcaption}
\usepackage{pifont}
\usepackage{stmaryrd}
\usepackage{multicol}

\usepackage{subcaption}
\usepackage{makecell}
\usepackage{adjustbox}
\usepackage{longtable}  
\usepackage{booktabs}   
\usepackage{mathastext}
\MTfamily{dejavusans}
\usepackage{graphicx}%
\usepackage{multirow}%
\usepackage{amsmath,amssymb,amsfonts}%
\usepackage{amsthm}%
\usepackage{mathrsfs}%
\usepackage[title]{appendix}%
\usepackage{pifont}%
\usepackage{textcomp}%
\usepackage{manyfoot}%
\usepackage{booktabs}%
\usepackage{algorithm}%
\usepackage{algorithmicx}%
\usepackage{algpseudocode}%
\usepackage{listings}%
\usepackage{mathpazo} 
\usepackage[justification=raggedright,singlelinecheck=false]{caption}

\title{Systematic Evaluation of Large Vision-Language Models for Surgical Artificial Intelligence}

\author{Anita Rau$^1$
\and
Mark Endo$^1$\\
\and
Josiah Aklilu$^1$ \\
\and
Jaewoo Heo$^1$\\
\and
Khaled Saab$^2$ \\
\and
Alberto Paderno$^3$\\
\and
Jeffrey Jopling$^4$ \\
\and
F. Christopher Holsinger$^1$ \\
\and
Serena Yeung-Levy$^1$ \\$ $ \\ $ $ \\ 
$^1$Stanford University \hspace{1em} $^2$Google DeepMind \hspace{1em} $^3$Humanitas University \\ $^4$Johns Hopkins University
\\ $ $ \\ 
\href{https://anitarau.github.io/surg-vlms-eval}{https://anitarau.github.io/surg-vlms-eval}
}

\begin{document}
\twocolumn[{%
\maketitle
\renewcommand\twocolumn[1][]{#1}%
\vspace{-1em}
\begin{abstract}

Large Vision-Language Models offer a new paradigm for AI-driven image understanding, enabling models to perform tasks without task-specific training. 
This flexibility holds particular promise across medicine, where expert-annotated data is scarce. Yet, VLMs’ practical utility in intervention-focused domains—especially surgery, where decision-making is subjective and clinical scenarios are variable—remains uncertain.
Here, we present a comprehensive analysis of 11 state-of-the-art VLMs across 17 key visual understanding tasks in surgical AI—from anatomy recognition to skill assessment—using 13 datasets spanning laparoscopic, robotic, and open procedures. In our experiments, VLMs demonstrate promising generalizability, at times outperforming supervised models when deployed outside their training setting. In-context learning, incorporating examples during testing, boosted performance up to three-fold, suggesting adaptability as a key strength. Still, tasks requiring spatial or temporal reasoning remained difficult. Beyond surgery, our findings offer insights into VLMs’ potential for tackling complex and dynamic scenarios in clinical and broader real-world applications.

\end{abstract}}]

Large Vision-Language Models (VLMs) are a new frontier in artificial intelligence for image and video understanding. These models acquire conceptual knowledge by linking images with descriptive, free-form text.  The learned associations allow these models to reason across modalities and interpret new visual inputs using the context and relationships learned from both images and text data. Like Large Language Models (LLMs), these models gain their generalizability from large-scale pretraining on unlabeled or weakly labeled data, allowing them to follow text-based instructions to tackle new problems in a ``zero-shot" setting---without requiring additional training, annotations, or bespoke AI systems. This advancement marks a departure from previous AI paradigms that relied on supervised models trained with human-annotated data to solve specific problems. This flexibility holds particular promise in medicine, where annotated data is often scarce and may not reflect the full range of real-world clinical scenarios. Already, VLMs have demonstrated remarkable capabilities in interpreting diverse biomedical imaging modalities, including microscopic, radiological, endoscopic, and natural images.  For example, they can classify pathology images \cite{huang2023visual}, identify intracardiac devices and evaluate cardiac function \cite{christensen2024vision}, detect abnormalities in CT chest scans \cite{hamamci2024foundation}, and retrieve dermatology images based on free-form textual descriptions \cite{kim2024transparent}---without task-specific training. Precisely, while these models may be trained on in-domain data for related tasks, they are not trained using labels specific to the evaluation or target tasks.

\begin{figure*}[t] 
    \caption*{\textbf{A}}
    \centering
    \begin{subfigure}{\textwidth}
        \centering
        \includegraphics[width=\textwidth, clip, trim=5em 21em 5em 21em]{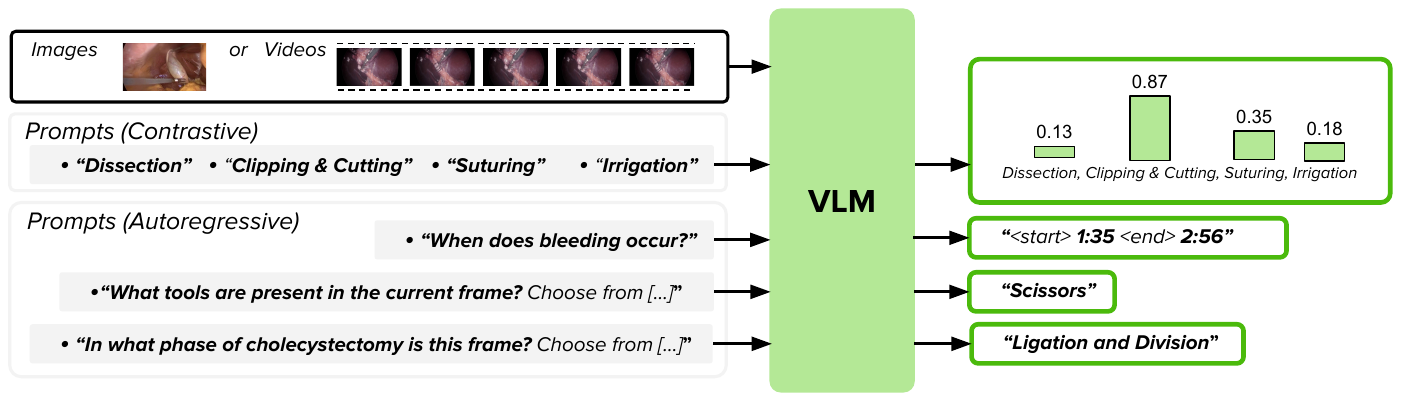}
        \label{fig:pull_figure}
    \end{subfigure}
    \vspace{-3em}
    \caption*{\textbf{B}}
    \begin{subfigure}{\textwidth}
        \centering
        \resizebox{\textwidth}{!}{%
        \begin{tabular}{ccccc}
\toprule Task & Surgical Training & OP Notes & Workflow & Augmentation \\
\midrule
  Recognizing \textbf{Tools, Hands, \& Anatomy} & 
\cellcolor[HTML]{F3F2F2} \textcolor[HTML]{666666}{\ding{52}} & 
\cellcolor[HTML]{FFE3E6} \textcolor[HTML]{E65CA0}{\ding{52}} & 
\cellcolor[HTML]{FFF6D9} \textcolor[HTML]{F1C231}{\ding{52}} & 
\cellcolor[HTML]{E0EFF6} \textcolor[HTML]{1CB7E9}{\ding{52}} \\
 Detecting \textbf{Tools, Hands, \& Anatomy} & 
\cellcolor[HTML]{F3F2F2} \textcolor[HTML]{666666}{\ding{52}} & 
\cellcolor[HTML]{FFE3E6} \textcolor[HTML]{E65CA0}{\ding{52}} & 
\cellcolor[HTML]{FFF6D9} \textcolor[HTML]{F1C231}{\ding{52}} & 
\cellcolor[HTML]{E0EFF6} \textcolor[HTML]{1CB7E9}{\ding{52}} \\
  Segmenting \textbf{Tools, Hands, \& Anatomy} & 
\cellcolor[HTML]{F3F2F2} \textcolor[HTML]{666666}{\ding{52}} & 
 & & 
\cellcolor[HTML]{E0EFF6} \textcolor[HTML]{1CB7E9}{\ding{52}} \\
\midrule
Recognizing \textbf{Phases} & & 
\cellcolor[HTML]{FFE3E6} \textcolor[HTML]{E65CA0}{\ding{52}} & \cellcolor[HTML]{FFF6D9} \textcolor[HTML]{F1C231}{\ding{52}} & \\

 Recognizing \textbf{Actions} & 
\cellcolor[HTML]{F3F2F2} \textcolor[HTML]{666666}{\ding{52}} & \cellcolor[HTML]{FFE3E6} \textcolor[HTML]{E65CA0}{\ding{52}} & \cellcolor[HTML]{FFF6D9} \textcolor[HTML]{F1C231}{\ding{52}} & 
\\

Recognizing \textbf{Gestures} & 
\cellcolor[HTML]{F3F2F2} \textcolor[HTML]{666666}{\ding{52}} & 
& & \\

\midrule
Assessing \textbf{Risk \& Safety} & 
\cellcolor[HTML]{F3F2F2} \textcolor[HTML]{666666}{\ding{52}} & \cellcolor[HTML]{FFE3E6} \textcolor[HTML]{E65CA0}{\ding{52}} &
& \\
Assessing \textbf{Skill} & 
\cellcolor[HTML]{F3F2F2} \textcolor[HTML]{666666}{\ding{52}} & 
& & \\
Recognizing \textbf{Errors} & 
\cellcolor[HTML]{F3F2F2} \textcolor[HTML]{666666}{\ding{52}} & \cellcolor[HTML]{FFE3E6} \textcolor[HTML]{E65CA0}{\ding{52}} & 
& 
\cellcolor[HTML]{E0EFF6} \textcolor[HTML]{1CB7E9}{\ding{52}} \\

\bottomrule
\end{tabular}
}
    \label{tab:pull_figure}
    \end{subfigure}
    \setlength{\fboxsep}{0pt}
    \caption{Overview of our evaluation framework. \textbf{A)} We prompt different contrastive and autoregressive VLMs with an instruction and an image or video. \textbf{B)} We identified four surgical applications that could be improved with AI assistance: {Surgical Training} aims to provide automated feedback and improve trainees' learning curves; {Augmentation} involves improving the surgeon’s view or offering intraoperative guidance; {OP Notes} generation aims to automate post-operative reporting to reduce surgeons' time commitment; Finally, {Workflow} addresses process efficiency to streamline surgical procedures and reduce wait times. We additionally identified proxy-tasks with existing public datasets that are foundational to these four broader surgical applications. We highlight in which surgical applications the identified technical tasks could be especially impactful. These tasks are at the center of our analysis.}
\label{fig:pull_figure_combined}
\end{figure*}

Despite their success in biomedical imaging, VLMs have yet to be widely applied to  intervention-focused areas of medicine, particularly surgery. Given surgery’s central role in healthcare, this gap is particularly striking. Each year, over 300 million surgeries are performed worldwide \cite{meara2015global} making surgery one of the most widely used medical interventions. But despite their prevalence, surgical procedures remain demanding and patient outcomes are heavily influenced by variability in surgical skill and individual preferences \cite{dencker2021postoperative}. AI-driven innovations in surgical training, workflow optimization, and intraoperative guidance could reduce variability in technique, assist in complex decision-making, prevent surgical errors, and ultimately improve patient safety.

Integrating these AI-driven advancements into surgical practice, however, presents unique challenges that differ from other medical domains. Unlike static imaging fields such as pathology and radiology, surgery involves dynamic, continuously changing scenes shaped by the movement of instruments and tissues.  Beyond anatomical variations, each surgeon employs a unique approach, resulting in rare cases that cannot be comprehensively captured in a dataset—the long tail of unseen conditions \cite{roy2022does,moor2023foundation}. Additionally, glossy textures, indistinct features, and obscuring elements like blood, fluids, smoke, occlusions, and inconsistent lighting further add to the complexity of the visual landscape. Finally, even when surgery is digitized, like with laproscopic and robotic surgery, the scarcity of saved recordings—often captured in ad-hoc setups—additionally limits the ability to train and fine-tune models.

While the scarcity of annotated, standardized, representative, and comprehensive data in surgery poses a significant challenge to training AI models, VLMs may be better equipped to address this limitation than traditional supervised methods. Through large-scale pre-training, VLMs can generalize to new domains that differ in appearance, context, or modality. VLMs even acquire zero-shot capabilities, enabling them to generalize to entirely new tasks \cite{radford2021learning,jia2021scaling,alayrac2022flamingo}. Where traditional supervised approaches require explicit and accurate image-label pairs to solve vision tasks, VLMs rely on large-scale language supervision---weak labels that are much easier to obtain than expert annotations for images. 

The ability to extrapolate learned concepts without additional training labels is driven by two primary VLM model architectures: contrastive models (e.g., CLIP \cite{radford2021learning}) and auto-regressive models (e.g., GPT-4o \cite{achiam2023gpt}). Contrastive models learn to associate images with their free-form captions, and at test time output how closely a new image is associated with a new caption. In the contrastive prompt in Figure \ref{fig:pull_figure_combined}A, the trained model computes this score between the query image and several possible captions. The final prediction is the caption leading to the highest score. Auto-regressive models, on the other hand, are directly based on LLMs which are trained to generate the next word fragment in a sequence. They primarily acquire conceptual knowledge from extensive text corpora (e.g., every book on the internet) minimizing the need for explicit human annotation. Afterwards, a separate visual encoder is trained to transform images into a format LLMs can understand. At test time, these models directly respond to queries, and there is no need to compute scores between images and caption candidates. Both approaches enable scalable pre-training, allowing VLMs to achieve their remarkable generalization capabilities and positioning them as potential candidates for surgical applications. Yet, VLMs' ability to handle the full complexity of real-world surgery remains uncertain and requires further exploration.

To assess whether VLMs can navigate the demands of surgery, we examine their capabilities across four key applications in AI-assisted surgery:  surgical training \cite{lam2022machine, aklilu2024artificial}, automated operative report generation \cite{berlet2022surgical,das2023automatic,khanna2023automated}, surgical field augmentation \cite{alfonso2020real,lungu2021review}, and workflow optimization \cite{franke2013intervention,twinanda2018rsdnet,mascagni2022computer,maier2017surgical}. These applications have been a consistent focus of researchers seeking to enhance surgery with AI assistance.  For each application, we identify different proxy tasks for which public evaluation data is available. For instance, to evaluate whether VLMs can automatically generate operative (OP) notes, we test their ability to recognize which tools are used on which anatomy. To assess whether a model has the potential to coach novice surgeons, we evaluate its accuracy at identifying skill and errors. Many of these technical tasks are shared between different downstream applications making them essential base capabilities for surgical VLMs. A mapping between the surgical applications of interest and the most relevant available proxy tasks is provided in Figure \ref{fig:pull_figure_combined}B. By utilizing public data, we aimed to establish an accessible evaluation framework that can serve as a benchmark for future VLMs in surgery. For critical tasks where no suitable public datasets exist, we collected private data. In summary, we provide a comprehensive overview of the current capabilities of VLMs in surgery and an evaluation framework for future models.
\section*{Results}\label{results}

\subsection*{Benchmarking framework for VLMs}
This study presents a comprehensive benchmarking framework for VLMs in surgical applications. We evaluate 11 VLMs, including eight autoregressive and three contrastive models. Among the autoregressive models, three are proprietary—GPT-4o \cite{hurst2024gpt}, Gemini 1.5 Pro \cite{team2024gemini}, and Med-Gemini \cite{saab2024capabilities}—while the others are openly available: Qwen2-VL \cite{wang2024qwen2}, PaliGemma 1 \cite{beyer2024paligemma}, LLaVA-NeXT \cite{liu2024llavanext}, InternVL 2.0 \cite{chen2024expanding}, and Phi-3.5 Vision \cite{abdin2024phi}. The contrastive models include CLIP \cite{radford2021learning}, OpenCLIP \cite{cherti2023reproducible}, and SurgVLP \cite{yuan2023learning}---the only available model specifically designed for surgical applications.

\begin{figure*}
    \begin{center}
    \includegraphics[width=0.9\linewidth, clip, trim=0.5em 0.5em 2em 0.5em]{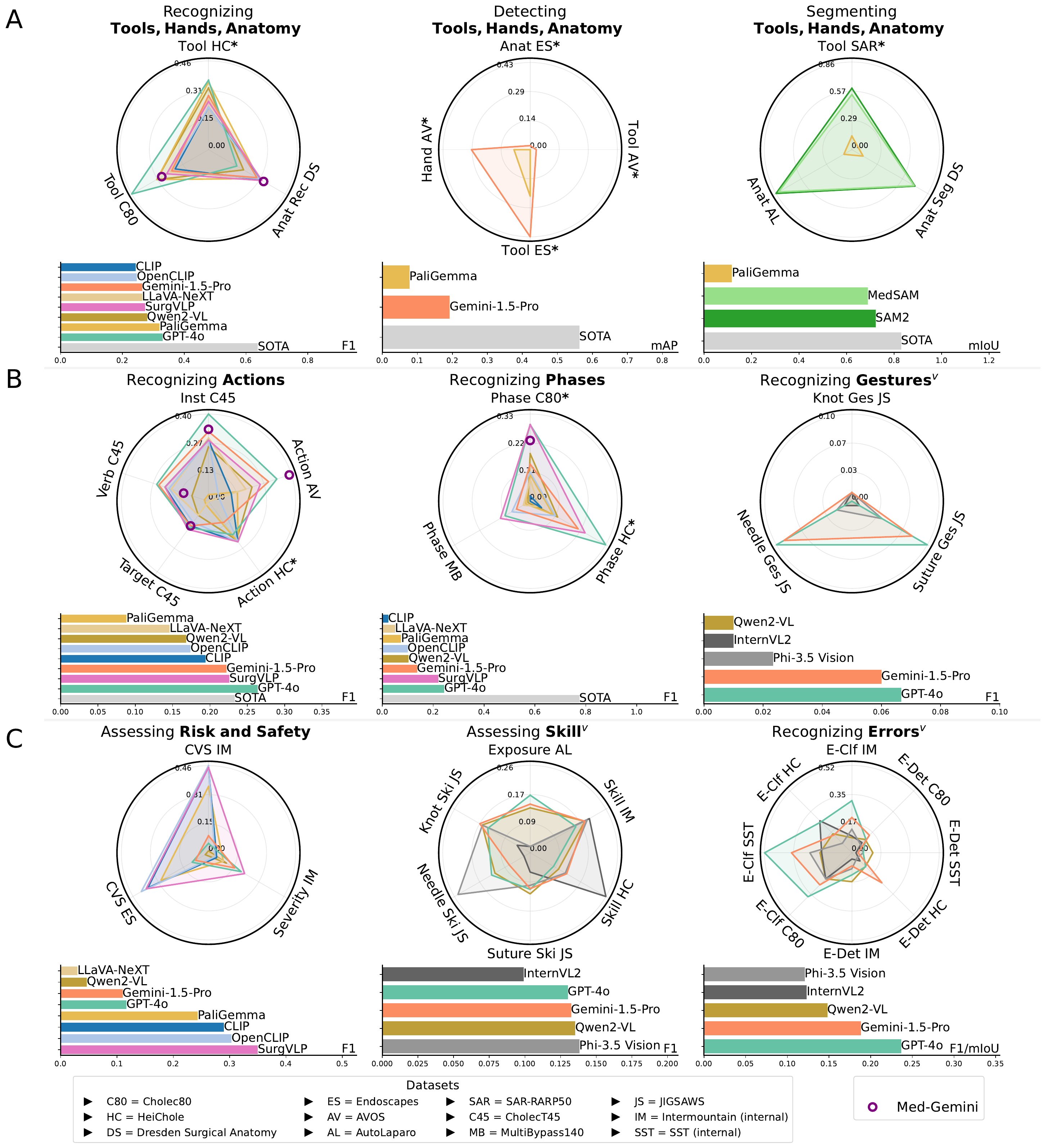}
    \end{center}
\caption{\small We evaluate 11 VLMs on 38 task instances. Larger values indicate better performance for all metrics. Bar plots compare average VLM performance to state-of-the-art supervised models (SOTA). SOTA values are averaged over official results where available (denoted by * and detailed in the Supplement). Med-Gemini results are sparse due to licensing restrictions. Task comparisons vary: Evaluating Gestures, Skill, and Errors are video-based tasks (denoted by $^v$) and require temporal understanding; detection and segmentation also require specialized spatial localization capabilities. Additional metrics in Supplement.     {\textbf{A) Surgical scene comprehension:} VLMs recognize surgical objects but struggle with localization; few support detection or segmentation.  To aid contextualization, we compare segmentation foundation models (SAM2/MedSAM), which generalize without training but are not VLMs. \textbf{B) Surgical progression understanding:} GPT-4o excels in procedural understanding, including action and phase recognition, with the open-source SurgVLP as a strong alternative. Gesture recognition in videos remains unsolved. \textbf{C) Surgical safety \& performance assessment:} Open-source contrastive models outperform proprietary ones in risk/safety assessment, and video tasks remain challenging. }}
    \label{fig:main}
\end{figure*}

Our evaluation framework spans 13 datasets, including 11 publicly available ones and spans open, laparoscopic, and robotic surgeries. Two additional private datasets were collected using the Black Box Explorer™ platform at Intermountain Health (IM) and anonymized partner sites of Surgical Safety Technologies (SST), comprising expert annotations for critical view of safety (CVS), surgical skill, and errors. Together, these datasets cover 17 visual tasks such as tool detection and phase recognition. Since many datasets support multiple tasks, our benchmark comprises 38 task-dataset pairs. For simplicity, we refer to them as \textit{task instances}. These instances are grouped into the nine tasks (Figs. \ref{fig:pull_figure_combined}B, \ref{fig:main}) and further categorized into three complexity levels: A) surgical scene comprehension including basic perception of objects in the surgical field; B) surgical progression understanding, including  understanding of the required steps in a surgery; and  C) surgical safety \& performance assessment including the ability to judge surgical competence and technical skills. While most tasks are image-based, our evaluation also includes video-tasks, namely gesture recognition, skill assessment, and error recognition.

Fig. \ref{fig:pull_figure_combined}A outlines our evaluation pipeline: each VLM is queried with a prompt and an input image or video, and its response is formatted for evaluation against ground truth annotations using F1 score, mAP (for detection tasks), or mIoU (for localization and segmentation tasks). Full details on datasets, tasks, models, prompts, and metrics are provided in the Supplementary Material.

\subsection*{Zero-shot performance across surgical tasks}
\begin{figure*}
    \begin{center}
    \includegraphics[width=0.94\linewidth]{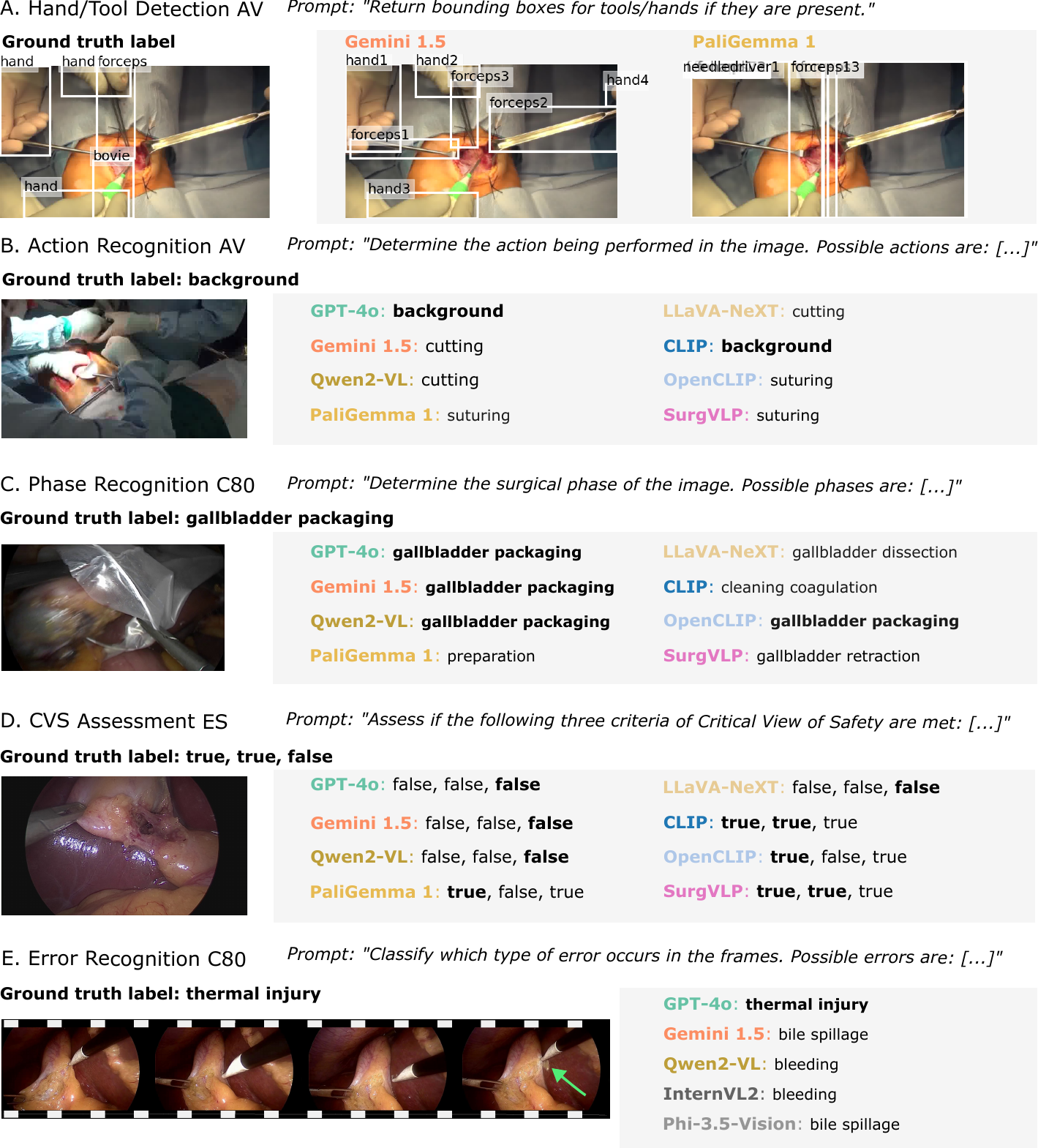}
    \end{center}
    \footnotesize{
    Datasets: AV=AVOS, C80=Cholec80, ES=Endoscapes}
    \caption{Qualitative zero-shot examples for various tasks, models, and datasets. Correct predictions are shown in \textbf{bold}. Prompts shortened for display; full versions in the Supplement. \textbf{A)} Gemini impresses in hand and tool detection, even spotting unannotated tools and hands. PaliGemma repeatedly predicts the same objects. \textbf{B)} GPT-4o leads at action prediction. \textbf{C)} Generalist models can infer surgical knowledge from general knowledge, in this example tying the term ``gallbladder packaging" to the easily discernible plastic bag in the shown example. \textbf{D)} CVS assessment is challenging for auto-regressive models. Contrastive models such as SurgVLP and CLIP can more accurately discriminate subtle visual cues in this task. \textbf{E)} GPT-4o outperforms all models at error classification. The error can be identified based on the thermal injury on the liver which we marked here with a {\textcolor[HTML]{51ED76}{green arrow}}, but the injury is missed by all models except GPT.}
    \label{fig:qualitative}
\end{figure*}
We evaluated the zero-shot capabilities of all models, and present quantitative results in Fig.~\ref{fig:main}, and qualitative examples in Fig. \ref{fig:qualitative} and supplementary Fig. \ref{fig:quali_supp}. For readability, we use shortened model names throughout; the exact model versions assessed in this study can be found in the Supplement. Since not all models are applicable to every task, the number of evaluated models varies. For example, the contrastive models we evaluated are not readily suited for video-based tasks. Further, as Med-Gemini is not publicly available, the model could only be evaluated on datasets with CC BY 4.0 licenses, or those for which we obtained explicit permission for this study.

\textbf{Proprietary VLMs lead in surgical scene comprehension and progression understanding, but face challenges in surgical safety \& performance assessment.}
Proprietary models—GPT-4o, Gemini, and Med-Gemini—showed strong performance across surgical image and video understanding tasks. Collectively, their strengths were most evident in surgical scene comprehension (A-level tasks) and surgical progression understanding (B-level tasks). They faced notable challenges in risk \& safety assessment and skill assessment, both C-level tasks, falling short against open models.

GPT-4o, a proprietary, generalist auto-regressive VLM, performed best overall. It excelled at tool recognition, surpassing the next-best model by 48\% on Cholec80 (Tool C80 in Fig. \ref{fig:main}A) and 3\% on HeiChole (Tool HC in Fig. \ref{fig:main}A). However, GPT remained well below the reported state-of-the-art (SOTA) benchmark on HeiChole trained in a supervised manner (\text{F}1 = 0.62 vs. 0.34).
GPT-4o also excelled in surgical progression understanding (B-level tasks), including action recognition (see example in Fig. \ref{fig:qualitative}B) and phase recognition (see example in Fig. \ref{fig:qualitative}C). Specifically, GPT-4o surpassed the second-best model, SurgVLP, by 17\% on average in action recognition (left bar plot in Fig. \ref{fig:main}B) and 11\% on average in phase recognition (center bar plot in Fig. \ref{fig:main}B). For gesture recognition, which required classifying a short video clip into one of 15 possible gestures, such as \textit{pulling suture with both hands}, GPT-4o outperformed all other evaluated models. Yet, all models struggled ($\text{F1} < 0.1$), emphasizing the task’s difficulty.
GPT's capabilities also extended to error recognition, which involves classifying a clip into one of four errors, such as \textit{thermal injury} (see example in Fig. \ref{fig:qualitative}E). The model achieved an $F1$ score of up to 0.52 on SST, outperforming all competitors by at least 44\% (E-Clf SST in Fig. \ref{fig:main}C).

Gemini and Med-Gemini also performed well. Gemini was one of only two models capable of detecting tools and hands, i.e. predicting the bounding box coordinates around an object (see example in Fig. \ref{fig:qualitative}A), and outperformed PaliGemma by a factor of roughly 2 on these tasks (Tool ES, Hand AV in Fig. \ref{fig:main}A). 
Med-Gemini was built upon Gemini and fine-tuned and specialized for medicine. Its training data included visual question-answer pairs from domains such as dermatology and radiology, clinician-written  long-form responses to medical questions, and summaries of medical notes. Med-Gemini achieved strong performance in anatomy recognition, outperforming all models by at least 3\% (Anat Rec DS in Fig. \ref{fig:main}A), and action recognition on the AVOS dataset, surpassing all models by at least 18\% (Action AV in Fig. \ref{fig:main}B).

However, proprietary models underperformed on most C-level task instances. GPT-4o and Gemini lagged in CVS and disease severity assessment (e.g., CVS IM, CVS ES, Severity IM in Fig. \ref{fig:main}C), with open models like SurgVLP, OpenCLIP, CLIP, and PaliGemma outperforming them by at least 160\% on average (left bar plot, Fig. \ref{fig:main}C). These tasks required reasoning over fine-grained visual cues, with descriptions such as: ``\textit{A carefully dissected }... \textit{presenting an unimpeded view of }..." These subtle and imprecisely worded differences may be difficult to convey through language alone. While VLMs can follow prompts to solve new tasks, they depend on a accurate visual description of unknown concepts. This could explain why GPT, and autoregressive models in general, struggled with these tasks.  In contrast, CLIP, OpenCLIP, and SurgVLP are explicitly trained to separate distinct categories within a shared embedding space, allowing them to learn robust image features to distinguish even subtle visual cues and perform better on visually complex tasks. PaliGemma notably outperformed all proprietary models in safety tasks but showed a consistent prediction bias in CVS criteria, possibly reflecting chance rather than real understanding (left bar plot in Fig. \ref{fig:main}C).
In skill assessment (rating on a scale from 1 to 5), GPT and Gemini were again outperformed—Qwen exceeded their performance by 6\% and 35\%, respectively (center bar plot in Fig \ref{fig:main}C). One exception was GPT’s accuracy on the Exposure AL task, predicting how the laparoscope should be moved to achieve appropriate exposure—i.e., a clear view of the surgical field.  GPT also struggled with error localization, which involves identifying the temporal start and end points of an error. While GPT was capable of recognizing errors in individual frames (E-Clf C80, E-Clf SST, E-Clf HC, E-Clf IM in Fig. \ref{fig:main}C), it lacked the spatial and temporal reasoning required to accurately pinpoint when the errors occurred in a video clip. In this task, Qwen outperformed GPT across multiple datasets (E-Det C80, E-Det SST, E-Det HC, E-Det IM in Fig. \ref{fig:main}C). But despite better performance, Qwen often produced the same prediction regardless of input, suggesting that its higher scores may not reflect capabilities. Additionally, GPT frequently returned exceptions for this task via API calls, with no clear cause.

\textbf{When specialization matters: SurgVLP dominates in tasks that require extensive surgical expertise.} 
SurgVLP, the only model in this study trained specifically for applications in surgery, consistently outperformed general-purpose VLMs in surgical tasks requiring expert knowledge. Pre-trained on video-caption pairs from various surgeries, it ranked highly in anatomy recognition (Anat Rec DS in Fig. \ref{fig:main}A) and phase recognition (Phase MB, Phase C80 in Fig. \ref{fig:main}B) and placed second overall in action recognition (left bar plot in Fig. \ref{fig:main}B). SurgVLP also achieved the highest performance in risk and safety assessment—a task requiring fine-grained visual interpretation in a clinical context. On average, SurgVLP surpassed all other evaluated methods by at least 15\% and outperformed GPT-4o by 200\% (left bar plot in Fig. \ref{fig:main}C).
Although SurgVLP was not explicitly trained to predict disease severity, it still demonstrated strong performance in this domain. The dataset used for disease severity benchmarking was private, annotated using a custom expert-defined protocol, and unlikely to have been encountered during model training. This made disease severity assessment a robust test of generalization and highlighted the advantage of domain-specific models in surgical contexts, particularly for complex clinical decision-making.

Despite these successes, SurgVLP ranked only fourth overall on tool, hands, and anatomy recognition (left bar plot in Fig. \ref{fig:main}A). However, this likely reflected the strength of competing models in tasks that do not require extensive surgical expertise, but the perception of generally known concepts like hands.
When compared with the SOTA performance for each task (gray bars in bar plots), SurgVLP came closest in action recognition but deviated most in phase recognition. These trends are consistent across all evaluated VLMs, suggesting that while domain-specific models offer advantages, performance gaps compared to task-specific SOTA models remain.

\textbf{Contrastive generalist VLMs struggle to generalize.} CLIP and OpenCLIP, both contrastive learning-based models, generally performed poorly across most surgical tasks. OpenCLIP is one of the few fully open-source models to release both its trained weights and training data.  Both models exhibited unexpected strength in action recognition, particularly in multi-label binary classification tasks (Action HC, Target C45, Verb C45, Inst C45 in Fig. \ref{fig:main}B). However, their performance dropped sharply in multi-class settings (Action AV in Fig. \ref{fig:main}B and Severity IM in Fig. \ref{fig:main}C), where they achieved F1 scores of just 0.08 and 0.05, respectively. This disparity likely stemmed from the evaluation metric. In multi-label binary classification settings, the predicted class was chosen based on the highest similarity to the evaluation image, whereas in binary classification, predictions depended on whether the similarity score surpassed a dataset-optimized threshold. This threshold-based approach effectively "learned" an optimal decision boundary from the dataset, potentially inflating performance in binary tasks without true generalization.
Notably, SurgVLP did not exhibit this discrepancy, maintaining consistent performance across both binary and multi-class tasks. 

\textbf{Open auto-regressive models have unpredictable strengths.} Open auto-regressive models exhibited inconsistent strengths across surgical tasks, with some models excelling in surprising areas while underperforming in others.
PaliGemma, despite weaker overall performance, stood out as the only model capable of image segmentation (Tool SAR, Anat AL, Anat Seg DS in Fig. \ref{fig:main}A). To better contextualize its performance, we compared it to two foundation segmentation models: MedSAM (fine-tuned for the mediBoth performed well on surgical tasks without further training, whereas PaliGemma struggled. Notably, SAM2 matched or outperformed MedSAM on both tool and anatomy segmentation tasks, suggesting that medical-specific fine-tuning is not always necessary. Meanwhile, PaliGemma exhibited substantially weaker segmentation capabilities, on average achieving only 16\% of SAM2's mIoU (left bar plot in Fig. \ref{fig:main}A). 
Outside segmentation, PaliGemma led among open auto-regressive models for anatomy and tool recognition (left bar plot in Fig. \ref{fig:main}A), and risk and safety assessment (left bar plot in Fig. \ref{fig:main}C), but ranked lowest in action recognition (left bar plot in Fig. \ref{fig:main}B), highlighting its unbalanced performance.

LLaVA-NeXT and InternVL2  generally struggled across tasks. In contrast,  Phi 3.5 Vision and Qwen2 performed surprisingly well in skill and error detection (center and right bar plots in Fig. \ref{fig:main}C)—both video-based tasks---but poorly in gesture recognition (right bar plot in Fig. \ref{fig:main}B), another video-based task.
These findings highlight the task-dependent nature of open auto-regressive models in surgical AI.

\textbf{Large-scale general training and small-scale domain-specific training perform comparable on surgical progression understanding tasks.} In surgical progression understanding tasks (B-level tasks), generalist models like GPT and Gemini performed similarly to the surgery-specific SurgVLP. In action recognition (left bar plot in Fig. \ref{fig:main}B) SurgVLP ranked between GPT and Gemini, while GPT-4o led overall but with varying strengths across tasks. For instance, GPT excelled in action triplet recognition (Inst C45, Verb C45, Target C45) whereas SurgVLP outperformed on Action HC.
 
Phase recognition showed a similar pattern. GPT and Gemini effectively identified distinct cholecystectomy phases (Phase C80, Phase HC in Fig. \ref{fig:main}B)—such as \textit{calot triangle dissection} (F1=0.72/0.43) and \textit{gallbladder packaging} (F1 = 0.53/0.73), likely by leveraging general knowledge (e.g. recognizing ``packaging" via retrieval bag as in Fig. \ref{fig:qualitative}D).
On gastric bypass surgery, however, GPT and Gemini performed significantly worse (Phase MB in Fig. \ref{fig:main}B), with F1 scores under 0.03 on both broad, e.g. \textit{preparation} and \textit{disassembling}, and niche phases, \textit{anastomosis test} and \textit{mesenteric defect closure}.  
SurgVLP, pre-trained on gastric bypass videos, achieved non-zero F1 scores across all phases, showing its domain-specific advantage.

\begin{figure*}
    \includegraphics[width=\linewidth]{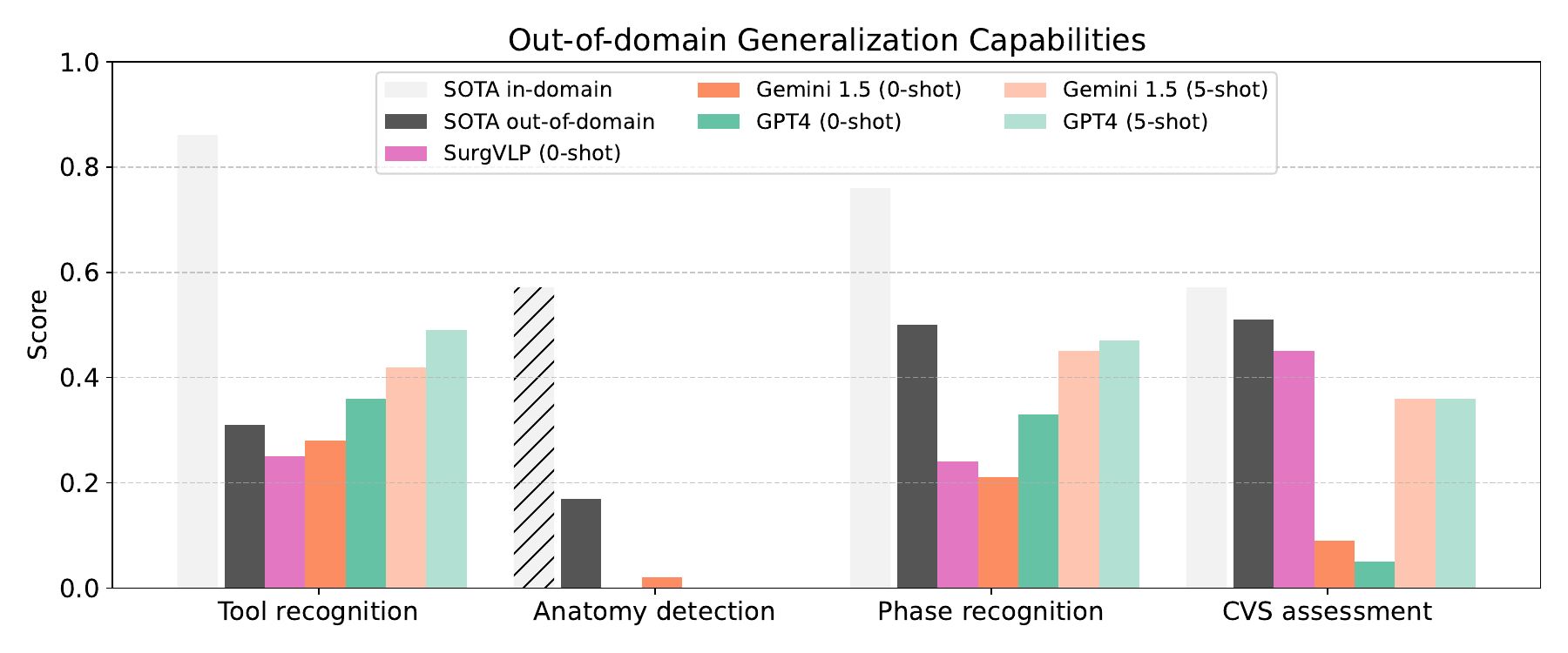}
    \caption{Comparing VLMs (colored bars) with task-specific SOTA models that are evaluated out-of-domain (dark gray) highlights the generalization capabilities of VLMs. In  this experiment SOTA models were evaluated on a different dataset than they had been trained on, but still performed the same task as during training. In this out-of-domain setting, VLMs perform competitively with SOTA models and even surpass them in zero-shot tool recognition. We also compare 5-shot results for models that have in-context capabilities (GPT and Gemini). Performance is reported using F1 scores for CVS assessment, phase recognition, and tool recognition, while anatomy detection is evaluated using mAP@.5:.95. For reference, we also include SOTA in-domain results, but note that these are not directly comparable to VLM results.}
    \label{fig:outofdomain}
\end{figure*}

\begin{figure*}
    \centering
    \includegraphics[width=\linewidth]{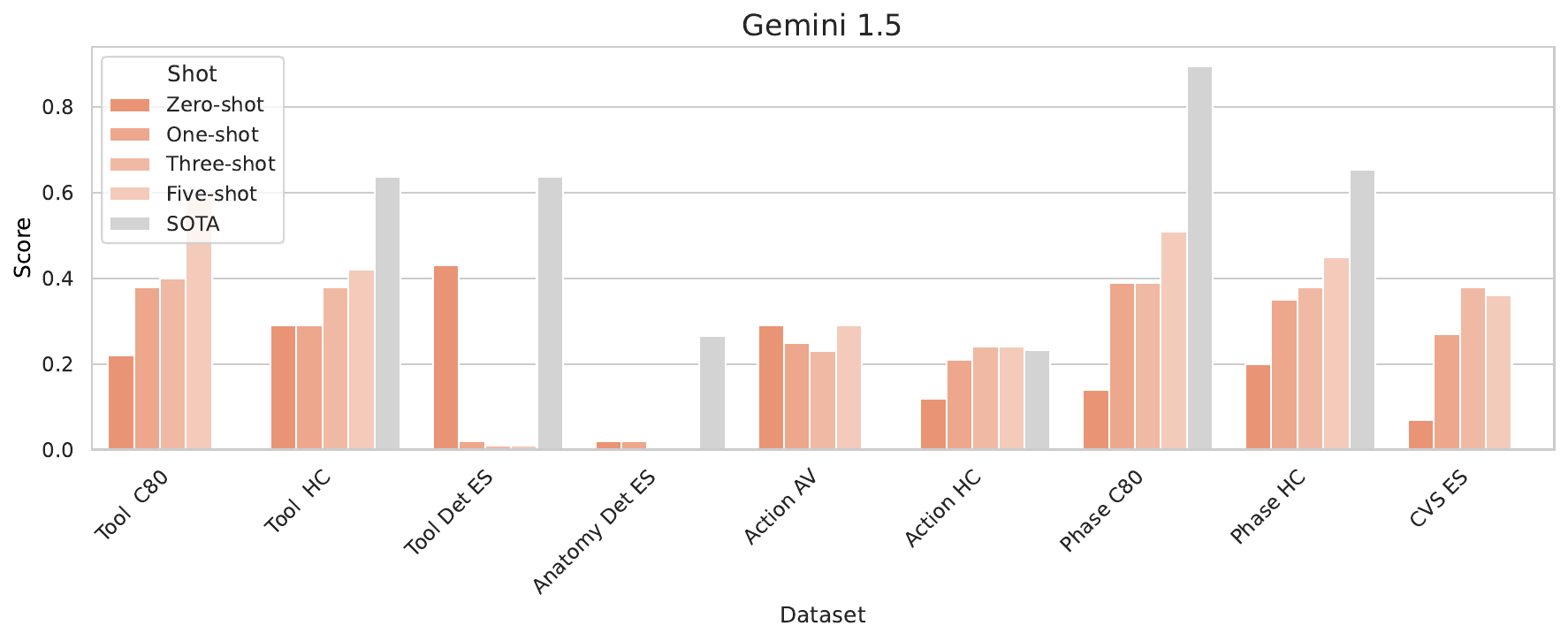}
    \includegraphics[width=\linewidth]{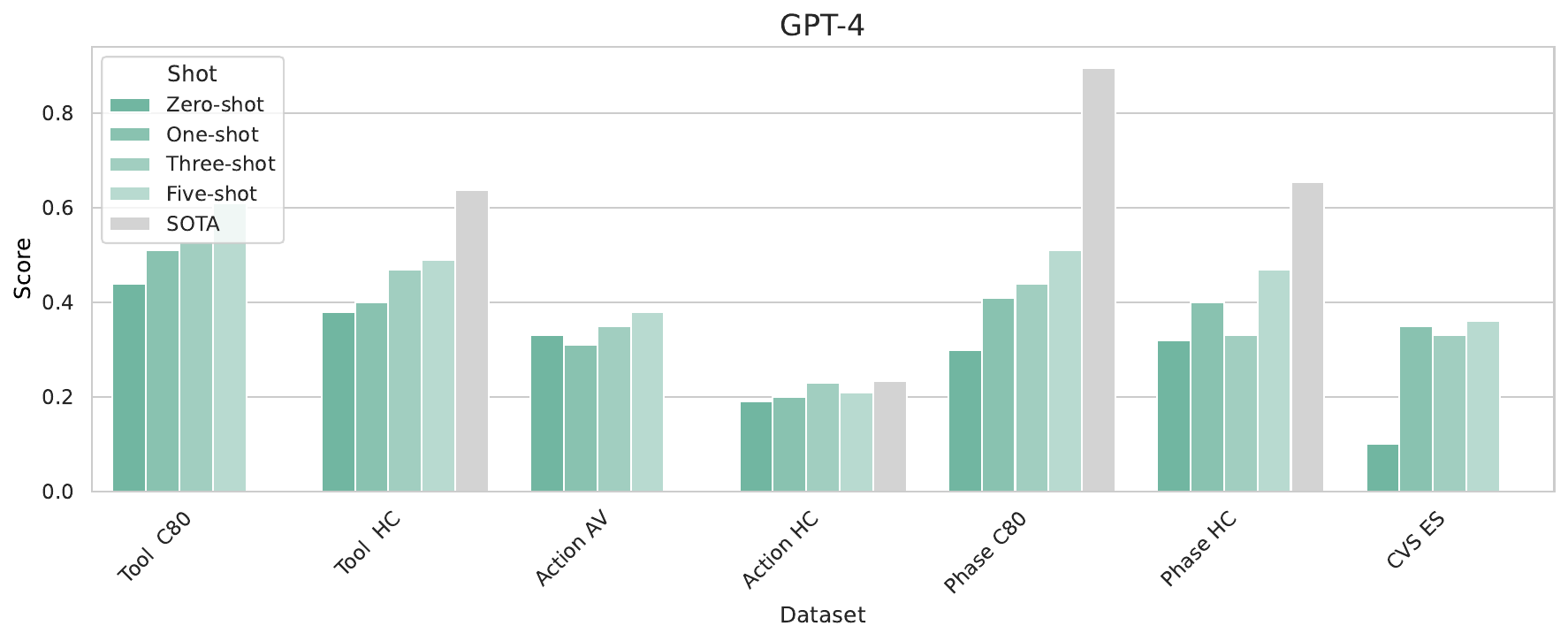}
    \caption{In-context learning by providing 1, 3, or 5 examples per class can improve model performance significantly versus the zero shot setting. Task-specific SOTA model results are provided for context when they are available. The score is F1 for all recognition tasks, and mAP@.5:.95 for detection (Det) tasks.}
    \label{fig:fewshot}
\end{figure*}
\subsection*{Comparison with out-of-domain state-of-the-art models}
In the previous section, we compared VLMs to task-specific SOTA models when evaluated in-domain for these SOTA models. For instance, for the tool recognition task on the HeiChole dataset (Tool HC Fig. \ref{fig:main}A), the task-specific SOTA model was trained to recognize tools on the training split of HeiChole, using provided training labels. The VLMs in this comparison were neither trained to recognize tools, nor trained on HeiChole data. As the two settings are not directly comparable, we investigated how the studied VLMs perform relative to task-specific models when the latter are evaluated out-of-domain—that is trained on one dataset and tested on another.

Figure~\ref{fig:outofdomain} compares the three leading VLMs, GPT, Gemini, and SurgVLP, to the out-of-domain SOTA results on four tasks for which two different datasets were available. The four tasks span all three task complexity levels (A-C). With this fairer comparison, we found the gap between VLMs and task-specific SOTA models shrank considerably. For tool presence, we even found that the performance of the state-of-the-art model dropped below that of GPT-4o in a zero-shot setting. In a five-shot setting, where VLMs with in-context capabilities (GPT-4o and Gemini) see five examples with labels at test time, VLMs exceed the performance of the SOTA model by up to 58\%. Details of the few-shot results are discussed in the next section. In phase recognition as well, 5-shot results almost match the task-specific model, and SurgVLP's zero-shot performance almost matches that of the SOTA model at CVS assessment. This out-of-domain comparison highlights the strong generalizability of VLMs compared to conventional, task-specific models. An exception is anatomy detection, where VLMs' performance is generally low and in-context learning does not improve results. Note that each task-specific model in Figure~\ref{fig:outofdomain} was explicitly trained for a single task and is limited to that task. Moreover, even under this fairer comparison, task-specific models only need to generalize to a new domain, whereas the evaluated VLMs must generalize to both a new domain and a new task.

\subsection*{In-context learning}

A key prospect of auto-regressive models achieving generalist medical artificial intelligence is the potential for in-context, or few-shot, learning \cite{moor2023foundation}, where models can learn from examples provided within the prompt without needing explicit training. Thus, we assessed whether in-context learning holds promise for the surgical tasks in our benchmark. For this study, we specifically focused on Gemini and GPT-4o which have in-context capabilities. As shown in Figure~\ref{fig:fewshot}, we found that for both Gemini  and GPT-4o, adding examples to the prompt indeed drastically improved model performance across a variety of tasks. In particular, for tasks such as CVS prediction, the F1 score of both GPT and Gemini increased more than three-fold. While GPT-4o, on average, outperformed Gemini, we found that Gemini especially benefited from in-context learning, with the F1 score of several task instances improving by at least two-fold, such as Action HC, CVS ES, Phase C80, Phase HC, and Tool C80. Although in-context learning showed promise across many surgical tasks, there were scenarios in which providing examples in the prompt either resulted in minimal performance increase or even degraded performance. Specifically, for tool detection, incorporating examples in the prompt caused performance to drop to nearly zero. One reason may be that the model does not interpret the example bounding boxes in the context of the task, and instead overfits to the provided values without understanding their spatial meaning. In summary, in-context learning shows promise as a direction for enhancing the performance of general-purpose models on surgical tasks, though it is not expected to always improve performance for any task.
\section*{Discussion}\label{discussion}
This study aimed to explore the current capabilities of state-of-the-art VLMs in surgery. To examine whether VLMs can serve as a generalist AI solution for surgery, we focused on four broad, clinically significant applications. For each application, we identified key technical tasks necessary to address them and systematically assessed the performance of a variety of state-of-the-art VLMs. Our findings suggest that while current models may not yet be deployment-ready in a zero-shot setting, their ability to generalize to new tasks and their adaptability in in-context learning scenarios highlights a promising avenue for further research.

To interpret the performance of VLMs, we evaluated them across several evaluation paradigms. In a zero-shot setting, the tested VLMs were able to tackle a variety of tasks spanning surgical scene understanding, surgical progression understanding, and surgical safety and performance assessment. But while the tested VLMs are versatile, they do not yet demonstrate the domain-specific understanding required for real-world deployment. This is expected given their lack of explicit surgical training, and highlights the fundamental challenge of applying generalist AI models to highly specialized tasks. Limited zero-shot performance is not unique to VLMs. Task-specific models also face serious limitations when tested outside their training distribution. While these models achieve high accuracy in expected, in-domain settings, their real-world applicability is constrained by the narrow conditions under which they are trained. The task-specific models we compare in this study usually involve testing methods on random subsets of a dataset, where test data—though from unseen patients—originates from the same hospital, under identical acquisition conditions, and with consistent annotators. This setup does not reflect real-world deployment, where AI systems must generalize across varied clinical environments. Instead, when evaluated under domain shift—such as data from different hospitals—the performance gap between task-specific models and VLMs diminishes. So, while VLMs may initially appear to underperform compared to task-specific models, a more realistic out-of-domain evaluation underscores their strong generalization capabilities relative to task-specific models. This does not mean that VLMs are deployment-ready. Rather, it highlights that neither the VLMs nor the task-specific models we evaluated in this study currently exhibit the level of robustness required for real-world surgical applications across the board. VLMs do offer viable paths forward, for instance, through in-context learning. 
 
VLMs' adaptability, particularly through in-context learning, makes them a promising direction for surgical AI.  With just a few examples, some auto-regressive VLMs show substantial performance gains over their zero-shot baseline, especially in recognizing surgical tools and phases.  On the CVS assessment task, a task requiring expert knowledge, the scale of improvement was especially impressive, suggesting that in-context learning allows VLMs to access such expert knowledge. When tested in out-of-domain settings, few-shot VLMs can approach or even surpass the performance of task-specific models trained on thousands of labeled images. This suggests that, unlike supervised models that require extensive training for specific tasks, VLMs can rapidly adapt to new surgical environments with minimal additional data.
This ability to rapidly adapt without retraining is particularly valuable in surgical domains where labeled data is scarce and difficult to curate. While much of surgical AI research has focused on routine procedures—where surgeries follow a predictable sequence of steps, large datasets are available, and complications are less frequent—real-world surgery is far more variable \cite{lavanchy2024challenges}. Many surgical sub-specialties lack the volume of annotated data needed to develop task-specific AI solutions. In these cases, VLMs’ ability to generalize across diverse tasks and learn from just a few examples makes them a uniquely scalable solution, offering AI-driven support in areas where traditional AI models have been more challenging to implement.
 
Despite promising generalization results, adaptability alone is not sufficient for clinical adoption, as key technical limitations persist. One major challenge is spatial reasoning—while many VLMs can identify the presence of surgical tools and anatomical structures, they struggle with more fine-grained localization tasks such as detecting bounding boxes. This weakness limits their utility in applications like surgical field augmentation, where precise object tracking is essential.
Another critical limitation is temporal reasoning---the ability to interpret the sequence and meaning of subtle changes and movements over time in video clips. Surgery is inherently a sequential process, requiring an understanding of how actions unfold over time. Without the ability to analyze fine-grained motions, their application to surgery remains limited.

These technical challenges have direct implications for the feasibility of different applications in AI-assisted surgery. Based on our analysis, the most promising near-term use cases for VLMs are workflow optimization and automated OP note generation. VLMs show encouraging performance in recognizing phases and identifying tool presence, both essential for workflow tracking and report automation. Since operative report generation relies heavily on text synthesis and structured reasoning—areas where VLMs already excel—this application appears to be a strong candidate for further research.
In contrast, surgical training remains a greater challenge. Gesture recognition, skill assessment, and error detection continue to be areas where VLMs have limitations. This is likely due to the limited temporal understanding capabilities of existing multi-modal models \cite{li2024vitatecs}. However, future advances in video-language models may improve temporal reasoning performance \cite{liu2024st}, potentially unlocking a range of applications in surgical training. 
For surgical field augmentation, the results present a mixed outlook. The strong segmentation performance of some foundation models suggests that AI could assist in highlighting anatomical structures and surgical landmarks, a key prerequisite for intraoperative guidance. However, VLMs’ inability to reliably detect errors remains a major limitation. Real-time augmentation requires an understanding of procedural deviations and unexpected events, a capability that current VLMs have yet to demonstrate.

While VLMs require further refinement before they can meaningfully impact surgical practice, their remarkable generalization capabilities make them a promising AI paradigm worthy of further exploration. To realize their full potential, researchers could focus on three key advancements. First, expanding pretraining datasets with high-quality surgical data—through international collaborations, multi-institutional efforts, or synthetic data augmentation—could significantly enhance domain-specific performance. Our results demonstrate that both large-scale pre-training with general data (e.g. GPT-4o) and small-scale fine-tuning on surgical data (e.g. SurgVLP), lead to strong model performance. A promising direction for future work is to combine these strengths by pre-training on large-scale unlabeled surgical data. This approach offers a scalable way to capture domain-specific knowledge without requiring expert annotations. 
The limited performance of general models on risk and safety assessment highlights their lack of semantic understanding of fine-grained surgical cues, a gap that surgical-domain pre-training could help close. Second, improving spatial and temporal reasoning remains a priority, as surgical AI models must accurately process procedural sequences and detect deviations. This need exists outside of the surgical domain, with current efforts improving temporal reasoning of multi-modal models across a variety of applications \cite{shangguan2024tomato}. A third critical consideration is the disparity between proprietary and open-source models—the best-performing VLMs in our study are proprietary, while open alternatives lag behind. And even among open models, most only provide access to their weights while withholding their training data. Although this performance gap is closing \cite{guo2025deepseek}, it raises concerns about accessibility, transparency, and reproducibility, all of which must be addressed to ensure trustworthy clinical integration.

With targeted advancements in reasoning capabilities and domain knowledge, VLMs could transition from experimental AI models to indispensable tools in surgical practice. 
Surgery presents a uniquely valuable challenge to develop such advancements, pushing models to interpret complex, high-variation environments in ways that could drive broader advancements in computer vision and multimodal learning. If these challenges are met, VLMs have the potential to reshape surgical AI—not as isolated task-specific tools, but as adaptable, multi-purpose systems. As the technology matures, understanding how and where VLMs can be safely leveraged will be key to ensuring their most effective use. Comprehensive and diverse benchmarking—across different surgical procedures, institutions, and applications—will be essential to accurately assess their performance, limitations, and readiness for integration into clinical practice. To this end, our benchmarking framework is publicly available and maintained at \href{https://anitarau.github.io/surg-vlms-eval}{https://anitarau.github.io/surg-vlms-eval}.

\textbf{Limitations:} While this benchmarking study highlights the broad capabilities of VLMs, several limitations remain. Our evaluation is constrained by available datasets, and while the tasks assessed provide valuable insights, they represent only isolated components of larger surgical applications. Further, as most models do not make their training data publicly available, it is possible that some of the VLMs we consider general models are exposed to some contaminated data---namely surgical data---making them partially in-domain VLMs. Ethical considerations, including bias, accountability, and the interpretability of model outputs, must also be addressed before VLMs can be broadly adopted in surgery. 

\subsection*{Acknowledgments}
This work was supported in part by the Isackson Family Foundation (C.H., A.R.), the Stanford Head and Neck Surgery Research Fund (C.H., A.R.), the Wellcome Leap SAVE program (No. 63447087-287892; S.Y., J.J., J.A., A.R.), the National Science Foundation (No. 2026498; S.Y.), and the National Science Foundation Graduate Research Fellowship Program (No. DGE-2146755; M.E.). Any opinions, findings, and conclusions or recommendations expressed in this material are those of the authors and do not necessarily reflect the views of any other entity.

\subsection*{Author Contributions}
\textit{Conceptualization}: A.R., S.Y., M.E.
\textit{Methodology}: A.R., S.Y., A.P, C.H, J.J.
\textit{Data Curation}: A.R.
\textit{Investigation}: A.R., M.E., J.A., J.H.; K.S. contributed to the benchmarking of Med-Gemini and was not involved in the evaluation of any other models.
\textit{Writing - Original Draft}: A.R., M.E., J.A., J.H.
\textit{Writing - Review \& Editing}: All authors.
\textit{Visualization}: J.A., A.R.
\textit{Supervision}: S.Y., C.H.
\textit{Funding acquisition}: S.Y., C.H., J.J.

\subsection*{Competing Interests}
K.S. is an employee of Alphabet and may own stock as part of the standard compensation package.

{
    \small
    \bibliographystyle{ieeenat_fullname}
    \bibliography{bibliography}
}

\section*{Methods}\label{methods}

\subsection*{Overview}
A detailed overview of all Vision-Language Models (VLMs), datasets, and tasks can be found in the Supplementary Material (Tables \ref{tab:vlms}, \ref{tab:datasets}, \ref{tab:tasks}). We chose a variety of models that either provide consistently high performance across diverse applications or are fundamental models that are often referenced in the literature. We also included the only surgery-specific VLM, SurgVLP, at the time of submission. For most tasks, we used the same prompt for all auto-regressive models; however, PaliGemma required a prompt tailored to its expected pattern. Contrastive models also required a specific prompt. For SurgVLP we followed the prompts used in the original paper as closely as possible. For the CVS task, we followed the prompt suggested in the Med-Gemini paper \cite{saab2024capabilities}. Prompts were tuned on the official validation sets, or training sets when no validation split was available. All prompts are included with the Supplementary Material (Table \ref{tab:prompts}).

Evaluation dataset were chosen based on a thorough review of existing public datasets. We used all datasets as they were intended by their authors. For instance, when videos were available, but the dataset was intended for frame-wise classification, we followed the original setup and predicted frame-wise labels. As many public datasets can only be used for research purposes, we were not able to obtain all results for Med-Gemini. 

Due to the high inference costs of commercial models, we limited test sets to approximately 10,000 test samples by reducing the frame rate of large dataset. Detailed frame rates per dataset can be found in the Supplement (Table \ref{tab:datasets}) .

For few-shot experiments, we randomly selected examples from the official training splits of the datasets. For multi-class binary classification problems, such as tool presence, we provided one image per class per shot. As some images have several true classes, this means that in our setting, more than one example per shot could be provided. We list the specific images and labels in the Supplementary Material (Table \ref{tab:few_prompts}). We subsampled the test sets for few-shot experiments by reducing the frame rate to limit the evaluation to approximately 1,000 images per dataset. For direct comparability, we repeated the zero-shot examples reported in Figure \ref{fig:main} on these smaller subsets for Figure \ref{fig:fewshot}. Frame rates per dataset are provided in the Supplement (Table \ref{tab:datasets}).

\subsection*{Non-VLM Generalization Experiments}
To systematically evaluate the generalization capabilities of state-of-the-art (SOTA) task-specific models in surgical applications, we trained each model on a domain-specific dataset and subsequently assessed its performance on an out-of-domain dataset. This evaluation was conducted across four distinct surgical tasks: (1) tool recognition, (2) phase recognition, (3) anatomy detection, and (4) critical view of safety (CVS) assessment. 

Each experiment followed a standardized approach where:
\begin{itemize}
    \item A SOTA model was trained on an in-domain dataset using the training protocol established in prior literature.
    \item The trained model was then evaluated on an out-of-domain dataset annotated for the same task.
    \item Performance metrics were reported to analyze the degradation in performance and to compare generalizability across different tasks.
\end{itemize}

This experimental design allows us to assess the generalization properties of task-specific models and directly compare them with general-purpose VLMs in their ability to extend beyond their training distribution.

\paragraph*{Tool Presence Prediction}
For tool presence prediction, we trained MoCo v2, a self-supervised contrastive learning model \cite{chen2020improvedbaselinesmomentumcontrastive}, on the Cholec80 dataset, which consists of 80 laparoscopic cholecystectomy videos collected in Strasbourg, France \cite{twinanda2016endonet}. The trained model was subsequently evaluated on the HeiChole dataset \cite{wagner2023comparative}, which contains laparoscopic cholecystectomy videos from University Hospital of Heidelberg, Germany, and its affiliate hospitals. Though there are discrepancies between Cholec80 and HeiChole's annotated tool class nomenclature, a 1-to-1 mapping is possible.  We referred to \cite{ramesh2023dissecting} for model definition, data-loading, and training \& evaluation pipeline.

\paragraph*{Phase Prediction}
For phase prediction, the MoCo v2 model was fine-tuned on Cholec80 and tested on HeiChole. As both datasets annotate the same phases they are directly comparable. By evaluating the trained model on HeiChole, we analyzed its capacity to recognize surgical phases in a new clinical setting, providing insights into the transferability of phase recognition models across institutions. 

\paragraph*{Anatomy Detection}
To evaluate generalization in anatomy detection, we trained a Faster-RCNN model on the CholecSeg8k dataset \cite{rios2023cholec80}, which consists of 8,080 laparoscopic frames extracted from Cholec80 and annotated at the pixel level for 13 anatomical structures. The model was then tested on Endoscapes \cite{murali2023endoscapes}, a dataset that includes laparoscopic images that, like Cholec80, where collected in Strasbourg, France. To ensure consistency, we focused on detecting the gallbladder, a commonly annotated structure in both datasets. This experiment allowed us to assess how well a model trained on one surgical anatomy dataset could generalize to another with different lighting conditions and camera perspectives. We referred to \cite{ramesh2023dissecting} for model definition, data-loading, and training \& evaluation pipeline.

\paragraph*{Critical View of Safety (CVS) Prediction}
For critical view of safety (CVS) prediction, we trained the LG-CVS \cite{murali2023latentgraphrepresentationscritical} model on the Endoscapes dataset and evaluated it on a private dataset collected in the United States. LG-CVS is a latent graph representation-based method that integrates object detection with structured anatomy-aware scene understanding \cite{murali2023latentgraphrepresentationscritical}. The private dataset was chosen to test the model's ability to recognize the critical anatomical structures required for safe laparoscopic cholecystectomy in a new surgical environment. This experiment provides valuable insights into how well structured graph-based representations can generalize across surgical settings.

\subsection*{Evaluation Metrics}
\paragraph*{Classification}
In this work, we report and compare three metrics: F1 score, weighted F1 score accounting for class imbalance, and accuracy. While we discuss the F1 score in the main paper, results on the other two metrics can be found in the Supplement (Tables \ref{tab:PhaseMB} - \ref{tab:E-DetIM}).  

Auto-regressive models do not provide confidence scores, so to evaluate them using metrics like mean average precision (mAP), one would have to assume a fixed confidence of 1 for all predictions. This assumption renders mAP comparisons with contrastive VLMs  or supervised classification models uninformative, which is why we do not report mAP for classification tasks.

Although accuracy is applicable to all model types, it is less reliable in our setting: many tasks involve presence classification with a large number of possible labels, most of which are negative. This leads to a highly imbalanced label distribution, where accuracy becomes overly optimistic due to the dominance of true negatives. For this reason, we adopt the F1 score as our primary evaluation metric.
To compute the F1 score for contrastive models, their similarity scores between images and prompts must first be converted into class labels. For multi-class classification tasks, we simply select the prompt with the highest similarity score as the predicted label. This approach works well and aligns naturally with the contrastive setup. For binary classification tasks, however, each prompt yields a single similarity score, necessitating the use of a threshold to convert scores into binary decisions.

To enable a fair comparison with auto-regressive models—especially under class imbalance—we compute the F1 score at optimal threshold (F1-max), following \cite{jeong2023winclip}. Specifically, we evaluate 200 thresholds uniformly spaced between the minimum and maximum similarity scores, as proposed in \cite{zou2022spot}. The threshold that maximizes the F1 score is selected independently for each class, and we report the average across all class-wise F1 scores.
It is important to note that this optimal threshold is determined post-hoc, meaning the resulting F1 score represents an upper bound on contrastive model performance under this evaluation setup.

\paragraph*{Segmentation and Object Detection}
We employ standard metrics for segmentation and object detection. For segmentation, we report the mean intersection over union (mIoU). For object detection, we report the mAP@[.5:.95] unless otherwise stated. For comparability with the state-of-the-art result, some tasks report the mAP@0.5. 

\subsection*{Private Datasets}
We collected several datasets to allow the evaluation of the tested VLMs on complex and interesting clinical tasks.

Our private Intermountain (IM) dataset \cite{aklilu2024artificial}, was collected in several hospitals affiliated with Intermountain Health, a multi-institutional not-for-profit healthcare system in the United States. The videos were collected during laparoscopic cholecystectomies by medical experts and included labels for:

\paragraph*{CVS} This dataset includes 3590 images each annotated with binary classification labels for three criteria. These include (1) Clear view of 2 tubular structures connected to the gallbladder; (2) A carefully dissected hepatocystic triangle presenting an unimpeded view of only the 2 cystic structures and the cystic plate; (3) The lower third of the gallbladder is dissected off the cystic plate. Each criterion is annotate as true or false, yielding three labels per image.
\paragraph*{Errors} This dataset includes 150 short video clips (180 seconds) of surgical errors are annotated as one of four classes: (1) Bleeding (blood flowing/moving from a source of injury that is clearly visible on the screen), (2) bile spillage (bile spilling out of the gallbladder or biliary ducts), (3) thermal injury (unintentional burn that leads to injury of non-target tissue), and (4) perforation (tool tissue interaction that leads to perforation of the gallbladder or biliary ducts and the spillage of bile). The beginning and end times of each error were then annotated for each clip. Each clip contains exactly one error.
\paragraph*{Skill} This dataset includes 74 short video clips annotated with surgical skill levels for five dimensions:
(1) Tissue Handling, (2) Psychomotor Skills, (3) Efficiency, (4) Dissection Quality, (5) Exposure Quality. Each skill type is rated by expert surgeons on a scale from 1 to 5. We measured inter-rater reliability between annotators to ensure annotation quality. In Figure \ref{fig:main}, we report the performance of a VLM averaged over the five skill types.

\paragraph*{Disease Severity} This dataset includes 68 images with annotations for disease severity (following the established Parkland Grading Scale for assessing cholecystitis) on a scale from 1 (less severe) to 5 (more severe): (1) Normal appearing gallbladder ("robins egg blue"), no adhesions present, completely normal gallbladder; (2) Minor adhesions at neck, otherwise normal gallbladder and adhesions restricted to the neck or lower of the gallbladder; (3) Presence of ANY of the following: hyperemia, pericholecystic fluid, adhesions to the body, distended gallbladder; (4) Presence of ANY of the following: adhesions obscuring majority of gallbladder, Grade 1-3 with abnormal liver anatomy, intrahepatic gallbladder, or impacted stone (Mirrizi); (5) Presence of ANY of the following: perforation, necrosis, inability to visualize the gallbladder due to adhesions. 
\\$ $ \\
We also collected video clips in collaboration with Surgical Safety Technologies (SST) who provided de-identified videos of laparoscopic cholecystectomy. While the location sites were not disclosed to us, the surgeries were performed at hospitals in the United States. This dataset was annotated with error labels by the same annotators as our Intermountain dataset.

\subsection*{Additional Details}
\paragraph{Reported SOTA results}
The reported SOTA results in Figure \ref{fig:main} are based on the following publications:
\begin{itemize}
    \item HeiChole results: Tool HC, Action HC, Phase HC \cite{wagner2023comparative}
    \item Cholec80 results: Phase C80 \cite{hirsch2023self}
    \item AVOC results: Hand AV, Tool AV \cite{goodman2024analyzing}
    \item Endoscapes results: Anat ES, Tool ES \cite{murali2023endoscapes}
    \item SAR-RARP50 results: Tool SAR \cite{psychogyios2023sar}
\end{itemize}

\paragraph*{Video Frame Sampling}
 For JIGSAWS and HeiChole skill assessment (Knot Ski JS, Needle Ski JS, Suture Ski JS, Skill HC), JIGSAWS gesture classification (Knot Ges JS, Suture Ges JS, Needle Ges JS), and AutoLaparo exposure assessment (Exposure AL), we utilized the Qwen2-VL frame sampler \cite{wang2024qwen2}. For Exposure AL, each five-second video was sampled at a rate of three frames per second to align with the methodology outlined in the original publication. For Skill HC consisting of lengthy videos, frames were sampled at a rate of 0.2 frames per second. For all error classification tasks (E-Clf HC, E-Clf SST, E-Clf C80, E-Clf IM), video clips were 30 seconds in length and 32 frames were uniformly sampled from each clip for input to all models (except Gemini 1.5 Pro, where we leverage the native Gemini API video sampling procedure). For error detection and all other tasks (Knot Ges JS, Suture Ges JS, Needle Ges JS, Knot Ski JS, Needle Ski JS, Suture Ski JS, Skill IM, E-Det HC, E-Det SST, E-Det C80, E-Det IM) the maximum number of frames was set to 70 if the model could process that context length; otherwise, the maximum was limited to 35 frames.

\paragraph*{Segmentation Foundation Models}
To evaluate the segmentation foundation models SAM2 and MedSAM, we used the ground truth labels to extract a tight bounding box around the region of interest. These bounding boxes are then used to prompt the segmentation models to segment the object in the foreground and return a binary segmentation mask. Finally, the segmentation performance was evaluated using IoU metrics, comparing each predicted binary mask against its corresponding ground truth mask.

\clearpage
\onecolumn
\appendix
\noindent\LARGE{\textbf{Supplement}}
\setcounter{table}{0}
\renewcommand{\thetable}{S\arabic{table}}
\setcounter{figure}{0}
\renewcommand{\thefigure}{S\arabic{figure}}
\setcounter{section}{0}
\normalsize
\section{Models}
All model versions are specified in Table \ref{tab:vlms}.

\begin{table*}[h!]
    \centering
    \caption{List of evaluated VLMs. Open-source models provide full public access to their weights, training code, and training data. Open-weights models make their weights publicly available but not their training data. We specify the Hugging Face (HF) or API version of each model.}
    \label{tab:vlms}
        \begin{tabular}{lcccc}
            \toprule
            Model Name & HF/API Version & Type & Access & Domain \\
            \midrule
            GPT-4o \cite{hurst2024gpt}& gpt-4o-2024-08-06 & Autoregressive & Commercial & General \\
            Gemini 1.5 Pro \cite{team2024gemini}& gemini-1.5-pro & Autoregressive & Commercial & General \\
            Med-Gemini \cite{saab2024capabilities} & - & Autoregressive & - & Medical \\
            Qwen2-VL\cite{wang2024qwen2} & Qwen2-VL-7B-Instruct & Autoregressive & Open-Weights & General \\
            PaliGemma 1 \cite{beyer2024paligemma}& paligemma-3b-mix-448 & Autoregressive & Open-Weights & General \\
            LLaVA-NeXT \cite{liu2024llavanext}& llava-v1.6-vicuna-7b-hf & Autoregressive & Open-Source & General \\
            InternVL 2.0 \cite{chen2024expanding}& InternVL2-8B & Autoregressive & Open-Weights & General \\
            Phi-3.5 Vision \cite{abdin2024phi}& Phi-3.5-vision-instruct & Autoregressive & Open-Weights & General \\
            CLIP \cite{radford2021learning}& clip-vit-base-patch32 & Contrastive & Open-Weights & General \\
            OpenCLIP \cite{cherti2023reproducible}& laion/CLIP-ViT-H-14-laion2B-s32B-b79K & Contrastive & Open-Source & General \\
            SurgVLP \cite{yuan2023learning}& - & Contrastive & Open-Weights & Surgical \\
            \bottomrule
        \end{tabular}
\end{table*}

\section{Datasets, Tasks, and Task-dataset Pairs}
This section includes an overview of all datasets (Table \ref{tab:datasets}) and tasks (Table \ref{tab:tasks}) used in this analysis. As some tasks exist in several datasets we also include an overview of all task-dataset combinations in Table \ref{tab:taskdataset}.

As some datasets have hundreds of thousand of test images, we subsample these extremely large datasets. Table \ref{tab:datasets} indicates in columns ``Zero-shot SR" and ``Few-shot SR" when test are subsampled and by which rate. As the context window of VLMs is limited, providing five examples for all classes is not always possible. We therefore subsample the test set in few-shot experiments more than in zero-shot experiments. For instance, for the HeiChole dataset, we only use every 375th frame during few-shot testing.

\begin{table*}[h!]
    \centering
    \caption{List of datasets used in this study.}
    \label{tab:datasets}
        \begin{tabular}{lccccc}
            \toprule
            Dataset & Surgery & Type  & Public & Zero-shot SR & Few-shot SR  \\
            \midrule
            AutoLaparo (AL) \cite{wang2022autolaparo}& L & I / V &   \ding{52}   & 1 &  -\\
            AVOS (AV) \cite{goodman2024analyzing}& O & I &   \ding{52} & 1 &  1\\
            Cholec80 (C80) \cite{twinanda2016endonet}& L & I &  \ding{52} & 5 & 15\\
            CholecT45 (C45) \cite{nwoye2022rendezvous}& L & I   &  \ding{52} & 1 & 1 \\
            Dresden Surgical Anatomy (DS) \cite{carstens2023dresden}& L & I   &  \ding{52}& 1 & - \\
            Endoscapes (ES) \cite{murali2023endoscapes}& L & I   &  \ding{52} & 1 & 2 \\
            HeiChole (HC) \cite{wagner2023comparative}& L & I / V  & \ding{52} &25 / 1 &375 / 1\\
            Intermountain (IM) & L & I / V   & \ding{55} & 1 & - \\
            JIGSAWS (JS) \cite{gao2014jhu}  & R & V   &  \ding{52} & 1 & - \\
            MultiBypass140 (MB) \cite{Lavanchy2024}& L & I   &  \ding{52} & 20 & - \\
            SAR-RARP50 (SAR) \cite{psychogyios2023sar} & R & I  & \ding{52} & 1 & - \\
            SST & L & V & \ding{55}  & 1 &  -\\
            \bottomrule
        \end{tabular}\\
    \footnotesize{SR=Subsample rate, I=Image, V=Video, L=Laparoscopic surgery, O=Open surgery, R=Robot-assisted surgery, M=Microsurgery.}
\end{table*}

\begin{table*}[h!]
    \caption{Overview of surgical tasks and associated datasets. Surgery types include laparoscopic (L), open (O), and robotic (R).}
    \label{tab:tasks} 
    \centering
    \setlength{\tabcolsep}{4pt} 
    {\footnotesize
        \begin{tabular}{l c >{\centering\arraybackslash}p{5.15cm} c}

            \toprule
            Task & Type & Datasets & Surgery Type \\
            \midrule
            Tool Presence & Image classification & Cholec80 \cite{twinanda2016endonet}, HeiChole \cite{wagner2023comparative} & L, L \\
            Anatomy Presence & Image classification & DresdenSA \cite{carstens2023dresden} & L \\
            Hand Detection & Bounding box estimation & AVOS \cite{goodman2024analyzing} & O \\
            Tool Detection & Bounding box estimation & Endoscapes \cite{murali2023endoscapes}, AVOS \cite{goodman2024analyzing} & L, O \\
            Anatomy Detection & Bounding box estimation & Endoscapes \cite{murali2023endoscapes} & L \\
            Tool Segmentation & Pixel-wise classification & SAR-RARP50 \cite{psychogyios2023sar} & R \\
            Anatomy Segmentation & Pixel-wise classification & AutoLaparo \cite{wang2022autolaparo}, DresdenSA \cite{carstens2023dresden} & L, L \\
            \midrule
            Action Recognition & Image classification & HeiChole \cite{wagner2023comparative}, AVOS \cite{goodman2024analyzing} & L, O \\
            Action Triplets & Image classification & CholecT45 \cite{nwoye2022rendezvous} & L \\
            Phase Recognition & Image classification & Cholec80 \cite{twinanda2016endonet}, MultiBypass140 \cite{Lavanchy2024}, HeiChole \cite{wagner2023comparative} & L, L, L \\
            Gesture Recognition & Video classification & JIGSAWS \cite{gao2014jhu} & R \\
            \midrule
            CVS Assessment & Image classification & Endoscapes \cite{murali2023endoscapes}, Intermountain & L, L \\
            Disease Severity & Video classification & \textit Intermountain & L \\
            Skill Assessment & Video classification & Intermountain, JIGSAWS \cite{gao2014jhu}, HeiChole \cite{wagner2023comparative} & L, R, L \\
            Exposure Assessment & Video classification & AutoLaparo \cite{wang2022autolaparo} & L \\
            Error Recognition & Video classification & Intermountain, SST, Cholec80 \cite{twinanda2016endonet}, HeiChole \cite{wagner2023comparative} (all own labels) & L, L, L, L \\
            Error Detection & Video segmentation & Intermountain, SST, Cholec80 \cite{twinanda2016endonet}, HeiChole \cite{wagner2023comparative} (all own labels) & L, L, L, L \\
            \bottomrule
        \end{tabular}}
\end{table*}

\begin{table*}[h!]
    \caption{List of task-dataset pairs. Number of samples corresponds to test set samples we used for evaluation. Classification types for classification tasks include multi-label binary classification (MLC) and multi-class classification (MCC).}
    \label{tab:taskdataset}
    \centering
    \setlength{\tabcolsep}{2pt} 
    {\footnotesize
    \begin{tabular}{lcccc}
        \toprule
        Pair & Classification Type & Super-Task & Task& \# Samples \\
        \midrule
        Tool C80 & MLC & \multirow{3}{*}{\makecell{Recognizing\\ Tools, Hands, Anatomy}} & Recognizing Tools & 15,367\\
        Tool HC & MLC &  & Recognizing Tools &14,259\\
        Anat Rec DS & MLC & & Recognizing Anatomy & 2,942 \\
        \midrule
        Tool ES & - & \multirow{4}{*}{\makecell{Detecting\\Tools, Hands, Anatomy}} &  Detecting Tools & 312\\
        Tool AV & - & & Detecting Tools & 2087\\
        Anat ES & - & & Detecting Anatomy & 312\\
        Hand AV & - & & Detecting Hands &2087\\
        \midrule
        Anat AL & - & \multirow{3}{*}{\makecell{Segmenting\\Tools, Hands, Anatomy}} & Segmenting Anatomy & 1,800\\
        Anat Seg DS & -& & Segmenting Anatomy & 13,195\\
        Tool SAR & - & & Segmenting Tools & 32,475\\
        \midrule
        Inst C45 & MLC & \multirow{5}{*}{\makecell{Recognizing Actions}} & Recognizing Action Triplets &3,823\\
        Verb C45 & MLC & & Recognizing Action Triplets &3,823\\
        Target C45 & MLC & &Recognizing Action Triplets &3,823 \\
        Action HC & MLC & & Recognizing Actions&14,259 \\
        Action AV & MCC & & Recognizing Actions & 292\\
        \midrule
        Phase C80 & MCC & \multirow{3}{*}{\makecell{Recognizing Phases}} & Recognizing Phases & 15,367\\
        Phase MB & MCC & & Recognizing Phases &11,542\\
        Phase HC & MCC & & Recognizing Phases & 14,259\\
        \midrule
        Knot Ges JS & MCC & \multirow{3}{*}{\makecell{Recognizing Gestures}} &Recognizing Gestures & 36\\
        Needle Ges JS & MCC & &Recognizing Gestures & 28\\
        Suture Ges JS & MCC & & Recognizing Gestures & 39\\
        \midrule
        CVS IM & MLC & \multirow{3}{*}{\makecell{Assessing\\Risk and Safety}} & Assessing CVS Achievement & 3,590\\
        CVS ES & MLC & & Assessing CVS Achievement & 1,799\\
        Severity IM & MCC & & Assessing Disease Severity & 68\\
        \midrule
        Exposure AL & MCC & \multirow{6}{*}{\makecell{Assessing Skill}} & Assessing Exposure & 73\\
        Skill HC & MCC & & Assessing Skill & 24\\
        Skill IM & MCC & & Assessing Skill & 74 \\
        Knot Ski JS & MCC & & Assessing Skill &  36\\
        Needle Ski JS & MCC & & Assessing Skill & 28 \\
        Suture Ski JS & MCC & & Assessing Skill & 39\\
        \midrule
        E-Clf IM & MCC & \multirow{8}{*}{\makecell{Recognizing Errors}} & Recognizing Errors & 150\\
        E-Clf HC & MCC & & Recognizing Errors & 50\\
        E-Clf SST & MCC & & Recognizing Errors & 140\\
        E-Clf C80 & MCC & & Recognizing Errors & 69\\
        E-Det IM & - & & Detecting Errors & 53 \\
        E-Det HC & - & & Detecting Errors & 9 \\
        E-Det SST & - & & Detecting Errors & 153 \\
        E-Det C80 & - & & Detecting Errors & 62 \\
        \bottomrule
    \end{tabular}}
\end{table*}

\clearpage 
\section{Prompts}
\renewcommand{\arraystretch}{1.3}  
{\footnotesize
\begin{longtable}[h!]{p{1.5cm} p{1.8cm} p{11cm}}
    \caption{List of prompts. } 
    \label{tab:prompts}\\
    \toprule
    Pair & Models & Prompt \\
    \midrule
    \endfirsthead

    \toprule
    Pair & Models & Prompt \\
    \midrule
    \endhead

    \midrule
    \multicolumn{3}{r}{\textit{Continued on next page...}} \\
    \midrule
    \endfoot

    \bottomrule
    \endlastfoot

\multirow{1}{*}{Tool C80} 
    & GPT, Gemini, Med-Gemini, Qwen2-VL, LLaVA-NeXT 
    & ``Which of these tools is present in the image: Grasper, Bipolar, Hook, Scissors, Clipper, Irrigator, SpecimenBag? Respond with a 0 or 1 for all tools according to whether or not the tool is present. Use this JSON schema: \{`tool\textunderscore{}name': bool\} and avoid line breaks." \\
    \cmidrule(lr){2-3}
    & PaliGemma & $[$``Answer en Is there a grasper in this image?", ``Is there a bipolar in this image?", ``Is there a hook in this image?", ``Are there scissors in this image?", ``Is there a clipper in this image?", ``Is there an irrigator in this image?", ``Is there a specimen bag in this image?"$]$ \\
    \cmidrule(lr){2-3}
    & CLIP, OpenCLIP & ``A surgical scene containing a $[$grasper, bipolar, hook, scissors, clipper, irrigator, specimen bag$]$". \\
    \cmidrule(lr){2-3}
    & SurgVLP & $[$``I use grasper or cautery forcep to grasp it", ``I use bipolar to coagulate and clean the bleeding", ``I use hook to dissect it", ``I use scissor", ``I use clipper to clip it", ``I use irrigator to suck it", ``I use specimenbag to wrap it"$]$ \\
    \midrule
    \multirow{1}{*}{Tool HC} 
    & GPT, Gemini, Qwen2-VL, LLaVA-NeXT & ``Which of these tools is present in the image: Grasper, Clipper, Coagulation instruments, Scissors, Suction-irrigation, Specimen bag, Stapler? 
        Respond with a 0 or 1 for all tools according to whether or not the tool is present. 
        Use this JSON schema: \{`tool\textunderscore{}name': bool\} and avoid line breaks." \\
    \cmidrule(lr){2-3}
    & PaliGemma & $[$``Answer en Is there a grasper in this image?", ``Is there a clipper in this image?", ``Is there a coagulation instrument in this image?", ``Are there scissors in this image?", ``Is there a suction-irrigation instrument in this image?", ``Is there a specimen bag in this image?", ``Is there a stapler in this image?"$]$ \\
    \cmidrule(lr){2-3}
    & CLIP, OpenCLIP & ``A surgical scene containing $[$a grasper, a clipper, coagulation instruments, scissors, suction-irrigation, a specimen bag, a stapler$]$."  \\
    \cmidrule(lr){2-3}
    & SurgVLP & $[$``I use grasper or cautery forcep to grasp it", ``I use clipper to clip it", ``I use bipolar to coagulate and clean the bleeding", ``I use scissor", ``I use irrigator to suck it", ``I use specimenbag to wrap it", ``I use stapler to staple it"$]$ \\
    \midrule
    \multirow{1}{*}{Anat Rec DS} 
    & GPT, Gemini, Med-Gemini, Qwen2-VL, LLaVA-NeXT & ``Which of these anatomical structures is visible in this image: abdominal wall, colon, inferior mesenteric artery, intestinal veins, liver, pancreas, small intestine, spleen, stomach, ureter, vesicular glands? Respond with a 0 or 1 for all structures according to whether or not the anatomy is visible. Use this JSON schema: \{`anatomy\textunderscore{}name': bool\} and avoid line breaks." \\
    \cmidrule(lr){2-3}
    & PaliGemma &$[$``Answer en Is the abdominal wall in this image?", ``Answer en Is the colon in this image?", ``Answer en Is the inferior mesenteric artery in this image?", ``Answer en Are intestinal veins in this image?", ``Answer en Is the liver in this image?", ``Answer en Is the pancreas in this image?", ``Answer en Is the small intestine in this image?", ``Answer en Is the spleen in this image?", ``Answer en Is the stomach in this image?", ``Answer en Is the ureter in this image?", ``Answer en Are vesicular glands in this image?"$]$ \\
    \cmidrule(lr){2-3}
    & CLIP, OpenCLIP & ``A surgical scene containing $[$the abdominal wall, the colon, the inferior mesenteric artery, intestinal veins, the liver, the pancreas, the small intestine, the spleen, the stomach, the ureter, vesicular glands$]$." \\
    \cmidrule(lr){2-3}
    & SurgVLP & $[$``I see the abdominal wall", ``I see the colon", ``I see the inferior mesenteric artery", ``I see intestinal veins", ``I see the liver", ``I see the pancreas", ``I see the small intestine", ``I see the spleen", ``I see the stomach", ``I see the ureter", ``I see vesicular glands"$]$ \\
    \midrule
    \multirow{1}{*}{\makecell[l]{Tool~ES \\Anat~ES}}
    & Gemini & 'Return bounding boxes for the cystic artery, cystic duct, cystic plate, gallbladder and all tools [ymin, xmin, ymax, xmax] if they are present. 
     Here are some examples: \{"cystic artery": [10,15,90,30], "tool1": [416, 96, 616, 406], "tool2": [654, 553, 959, 819]\} or \{"tool1": [416, 96, 616, 406]\} or \{ \}.'\footnote{Formatting examples, not few-shot examples.}  \\
    \cmidrule(lr){2-3}
    & PaliGemma & ['detect cystic artery', 'detect cystic duct', 'detect cystic plate', 'detect gallbladder', 'detect tool'] \\
    \midrule
    \multirow{1}{*}{\makecell[l]{Tool~AV \\Hand~AV}}
    & Gemini & 'Return bounding boxes for the bovies, forceps, hands, and needledrivers [ymin, xmin, ymax, xmax] if they are present.
      Here are some examples: \{"bovie1": [10,15,90,30], "hand1": [416, 96, 616, 406], "hand2": [654, 553, 959, 819]\} or \{"hand1": [416, 96, 616, 406]\} or \{ \}. Use the same output json format as in the examples.' \\
    \cmidrule(lr){2-3}
    & PaliGemma & ['detect bovie', 'detect forceps', 'detect hand', 'detect needledriver'] \\
    \midrule
        \multirow{1}{*}{Anat AL} 
    & PaliGemma & "\textless image\textgreater \textless bos\textgreater segment the uterus in the surgical image"\\
\midrule
        \multirow{1}{*}{Anat Seg DS} 
    & PaliGemma & "\textless image\textgreater \textless bos\textgreater segment the [abdominal wall, colon, inferior mesenteric artery, intestinal veins, liver, pancreas, small intestine, spleen, stomach, ureter, vesicular glands] in the surgical image"\\
\midrule
        \multirow{1}{*}{Tool SAR} 
    & PaliGemma & "\textless image\textgreater \textless bos\textgreater segment the [tool clasper, tool wrist, tool shaft, suturing needle, thread, suction tool, needle holder, clamps, catheter] in the surgical image"\\
\midrule
\multirow{1}{*}{\makecell[l]{Inst~C45\\Verb~C45\\Target~C45}}
    & GPT, Gemini, Med-Gemini, Qwen2-VL, LLaVA-NeXT & ``Find all instruments in these images of laparoscopic cholecystectomies. For each instrument, provide the action                                 it is performing and the tissue it is performing the action on. The following instruments are possible:                                 {grasper, bipolar, hook, scissors, clipper, irrigator, null}. Choose one of the following actions: {grasp, retract, dissect, coagulate,                                 clip, cut, aspirate, irrigate, pack, null}. Choose one of the following tissues: {gallbladder, cystic plate, cystic duct,                                 cysic artery, cystic pedicle, blood vessel, fluid, abdominal wall cavity, liver, adhesion, omentum, peritoneum, gut, specimen bag, null}.                                 Return a  dict using this JSON schema: \{"instrument": [tool1,...], "verb": [activity1,...], "target": [tissue1,...]\},                                 and avoid line breaks. If no instrument is present, return \{"instrument": ["null"], "verb": ["null"], "target": ["null"]\}. If an                                 instrument is present but no activity or tissue is visible, return \{"instrument": ["tool1", ...], "verb": ["null"], "target": ["null"]\}." \\
    \cmidrule(lr){2-3}
    & PaliGemma & ['Answer en Is there a grasper in this image?', 'Is there a bipolar in this image?', 'Is there a hook in this image?', 'Are there scissors in this image?', 'Is there a clipper in this image?', 'Is there a irrigator in this image?', 'Is there a specimen bag in this image?'] \\
    \cmidrule(lr){2-3}
    & CLIP, OpenCLIP & "A [list of instruments] is [list of verbs]ing the [list of targets]."  \textit{generate all possible combinations.}\\
    \cmidrule(lr){2-3}
    & SurgVLP & "I use a [list of instruments] to [list of verbs] the [list of targets]." \textit{generate all possible combinations.}\\
    \midrule

\multirow{1}{*}{Action HC} 
    & GPT, Gemini, Qwen2-VL, LLaVA-NeXT & ``You are shown an image captured during a laparoscopic cholecystectomy. Find all tools. For each tool, decide if the tool performs one of the following actions:     grasp, hold, cut, or clip. It is possible that the insturment is idle and no action is performed. Aggregate the actions across all instruments. For each action return a boolean indicating if the action is performed by any instrument.     Use this JSON schema: {"action": bool} and avoid line breaks. An example output could look like this: \{"grasp": bool, "hold": bool, "cut": bool, "clip": bool\}." \\
    \cmidrule(lr){2-3}
    & PaliGemma & ['Answer en Does one of the depicted tools grasp something?', 'Answer en Does one of the depicted tools hold something?', 'Answer en Does one of the depicted tools perform cutting on something?', 'Answer en Does one of the depicted tools clip something?'] \\
    \cmidrule(lr){2-3}
    & CLIP, OpenCLIP & ['A surgical tool grasping something', 'A surgical tool holding something', 'A surgical tool cutting something', 'A surgical tool clipping something'] \\
    \cmidrule(lr){2-3}
    & SurgVLP & ['I grasp it', 'I hold it', 'I cut it', 'I clip it'] \\
    \midrule
\multirow{1}{*}{Action AV} 
    & GPT, Gemini, Med-Gemini, Qwen2-VL, LLaVA-NeXT & ``You are shown an image captured during an open surgery. Determine the action being performed in the image. The possible actions are 0: cutting - if the surgeon is using a tool like scissors, scalpel, knife, or electrocautery device to cut or dissect tissues. 1: tying - if the surgeon is using their hands or needle holders to create secure knots. 2: suturing - if the surgeon is closing an open wound with a needle but not creating secure knots. 3: background - if no surgical action is being performed, including actions like using forceps, clamps, retractors, or dilators. There are no other options. Use this JSON schema: \{"action": int\} and avoid line breaks."  \\
    \cmidrule(lr){2-3}
    & PaliGemma & ``Answer en What is the surgical action being performed in this image? Choose from: cutting, tying knots, suturing, background task." \\
    \cmidrule(lr){2-3}
    & CLIP, OpenCLIP & ['A surgeon is cutting tissue with scissors, scalpel, knife, or electrocautery device', 'A surgeon is tying knots with their hands or needle holders', 'A surgeon is suturing an open wound creating straight stitches', 'A surgeon performs a background task like using forceps, clamps, retractors, or dilators'] \\
    \cmidrule(lr){2-3}
    & SurgVLP &['I cut it', 'I tie it', 'I suture it', 'I perform a background task'] \\
    \midrule
\multirow{1}{*}{\makecell[l]{Phase C80\\ Phase HC}} 
    & GPT, Gemini, Med-Gemini, Qwen2-VL, LLaVA-NeXT & ``You are shown an image captured during a laparoscopic cholecystectomy. Determine the surgical phase of the image. The possible phases are     0: Preparation, 1: Calot Triangle Dissection, 2: Clipping Cutting, 3: Gallbladder Dissection,     4: Gallbladder Packaging, 5: Cleaning Coagulation, 6: Gallbladder Retraction. There are no other options. Use this JSON schema: \{"phase": int\} and avoid line breaks."  \\
    \cmidrule(lr){2-3}
    & PaliGemma & ``Answer en What is the surgical phase shown in this image? Choose from: Preparation, Calot Triangle Dissection, Clipping Cutting, Gallbladder Dissection, Gallbladder Packaging, Cleaning Coagulation, Gallbladder Retraction." \\
    \cmidrule(lr){2-3}
    & CLIP, OpenCLIP & ``A surgical scene during [preparation, calot triangle dissection, clipping cutting, gallbladder dissection, gallbladder packaging, cleaning coagulation, gallbladder retraction]." \\
    \cmidrule(lr){2-3}
    & SurgVLP & ['In preparation phase I insert trocars to patient abdomen cavity', 'In calot triangle dissection phase I use grasper to hold gallbladder and use hook to expose the hepatic triangle area and cystic duct and cystic artery', 'In clip and cut phase I use clipper to clip the cystic duct and artery then use scissor to cut them', 'In dissection phase I use the hook to dissect the connective tissue between gallbladder and liver', 'In packaging phase I put the gallbladder into the specimen bag', 'In clean and coagulation phase I use suction and irrigation to clear the surgical field and coagulate bleeding vessels', 'In retraction phase I grasp the specimen bag and remove it from trocar'] \\
    \midrule
\multirow{1}{*}{Phase MB} 
    & GPT, Gemini, Qwen2-VL, LLaVA-NeXT & ``You are shown an image captured during a laparoscopic gastric bypass surgery. Determine the surgical phase of the image. The possible phases are     0: Preparation, 1: Gastric pouch creation, 2: Omentum division, 3: Gastrojejunal anastomosis, 4: Anastomosis test, 5: Jejunal separation, 6:Petersen space closure, 7: Jejunojejunal anastomosis,         8: Mesenteric defect closure, 9: Cleaning \& Coagulation, 10: Disassembling, 11: Other intervention. There are no other options. Use this JSON schema: \{"phase": int\} and avoid line breaks." \\
    \cmidrule(lr){2-3}
    & PaliGemma & ``Answer en What is the surgical phase depicted in this image? Choose one answer from this list: Preparation, Gastric Pouch Creation, Omentum Division, Gastrojejunal Anastomosis, Anastomosis Test, Jejunal Separation, Petersen Space Closure, Jejunojejunal Anastomosis, Mesenteric Defect Closure, Cleaning and Coagulation, Disassembling, Other Intervention." \\
    \cmidrule(lr){2-3}
    & CLIP, OpenCLIP & ``A surgical scene during [preparation, gastric pouch creation, omentum division, gastrojejunal anastomosis, anastomosis test, jejunal separation, petersen space closure, jejunojejunal anastomosis, 
        mesenteric defect closure, cleaning \& coagulation, disassembling, other intervention]."\\
    \cmidrule(lr){2-3}
    & SurgVLP & ['In preparation phase I insert trocars to patient abdomen cavity and prepare the surgical instruments', 'In gastric pouch creation phase I use stapler to create a small gastric pouch from the stomach', 'In omentum division phase I divide the omentum to prepare the space for the bypass', 'In gastrojejunal anastomosis phase I connect the gastric pouch to the jejunum using a stapler or sutures', 'In anastomosis test phase I test the anastomosis for leaks by injecting saline and observing for any leakage', 'In jejunal separation phase I use stapler to separate the jejunum at the appropriate length for the bypass', 'In petersen space closure phase I close the Petersen space to prevent internal hernia formation', 'In jejunojejunal anastomosis phase I connect the proximal and distal parts of the jejunum to ensure intestinal continuity', 'In mesenteric defect closure phase I close the mesenteric defect to prevent internal hernias', 'In cleaning \& coagulation phase I use suction and irrigation to clear the surgical field and coagulate bleeding vessels', 'In disassembling phase I remove the surgical instruments and prepare to close the abdomen', 'In other intervention phase I address any additional surgical requirements or complications as needed'] \\
    \midrule
\multirow{1}{*}{\makecell[l]{Knot Ges JS \\Needle Ges JS\\Suture Ges JS}}
    & GPT, Gemini, Phi-3.5-Vision, InternVL2 & ``You are a helpful medical video assistant. You will be provided with separate frames uniformaly sampled from a video segment. Task: classify the gesture of the surgical activity video segment. Below are the defined gestures: G1 Reaching for needle with right hand; G2 Positioning needle; G3 Pushing needle through tissue; G4 Transferring needle from left to right; G5 Moving to center with needle in grip; G6 Pulling suture with left hand; G7 Pulling suture with right hand; G8 Orienting needle; G9 Using right hand to help tighten suture; G10 Loosening more suture; G11 Dropping suture at end and moving to end points; G12 Reaching for needle with left hand; G13 Making C loop around right hand; G14 Reaching for suture with right hand; G15 Pulling suture with both hands.; Instructions: Assess the images carefully and classify the gesture. The segment only contains one gesture. Only output the gesture, eg: G1." \\
    \cmidrule(lr){2-3}
    & Qwen2-VL & ``You are a helpful medical video assistant. Task: classify the gesture of the surgical activity video segment. Below are the defined gestures: G1 Reaching for needle with right hand; G2 Positioning needle; G3 Pushing needle through tissue; G4 Transferring needle from left to right; G5 Moving to center with needle in grip; G6 Pulling suture with left hand; G7 Pulling suture with right hand; G8 Orienting needle; G9 Using right hand to help tighten suture; G10 Loosening more suture; G11 Dropping suture at end and moving to end points; G12 Reaching for needle with left hand; G13 Making C loop around right hand; G14 Reaching for suture with right hand; G15 Pulling suture with both hands.; Instructions: Assess the video segment carefully and classify the gesture. The segment only contains one gesture. Only output the gesture, eg: G1" \\
    \midrule

\multirow{1}{*}{\makecell[l]{CVS IM \\CVS ES}}
    & GPT, Gemini, Qwen2-VL, LLaVA-NeXT & ``You are a helpful medical video assistant. Task: Assess whether Critical View of Safety (CVS) is fully achieved in the provided frames from a cholecystectomy video. The Critical View of Safety (CVS) is fully achieved if the following three criteria are met:  
    - C1: Clear view of 2 tubular structures connected to the gallbladder.  
    - C2: A carefully dissected hepatocystic triangle presenting an unimpeded view of only the 2 cystic structures and the cystic plate.  
    - C3: The lower third of the gallbladder is dissected off the cystic plate.  

    Instructions: Assess the image carefully, and answer which of the Critical View of Safety (CVS) criteria are met. Use this JSON schema: \{`criterion': bool\} and avoid line breaks." \\
    \cmidrule(lr){2-3}
    & PaliGemma & $[$``Answer en Is there a clear view of 2 tubular structures connected to the gallbladder?", ``Answer en Is there a carefully dissected hepatocystic triangle presenting an unimpeded view of only the 2 cystic structures and the cystic plate?", ``Answer en Is the lower third of the gallbladder dissected off the cystic plate?"$]$ \\
    \cmidrule(lr){2-3}
    & CLIP, OpenCLIP & $[$``Clear view of 2 tubular structures connected to the gallbladder", ``A carefully dissected hepatocystic triangle presenting an unimpeded view of only the 2 cystic structures and the cystic plate", ``The lower third of the gallbladder is dissected off the cystic plate."$]$ \\
    \cmidrule(lr){2-3}
    & SurgVLP & $[$`2 tubular structures are connected to the gallbladder", ``A carefully dissected hepatocystic triangle is presenting an unimpeded view of only the 2 cystic structures and the cystic plate", ``The lower third of the gallbladder is dissected off the cystic plate."$]$ \\
    \midrule

\multirow{1}{*}{Severity IM} 
    & GPT, Gemini, Qwen2-VL, LLaVA-NeXT & ``What is the severity of inflammation in the provided image of the gallbladder in the initial stage
of surgery from a cholecystectomy video. The severity levels are from 1-5 according to the Parkland Grading Scale, where 
1 is least severe and 5 is most severe. The severity levels are as follows: 
1: Normal appearing gallbladder (`robins egg blue'), no adhesions present, completely normal gallbladder. 
2: Minor adhesions at neck, otherwise normal gallbladder. Adhesions restricted to the neck or lower of the gallbladder. 
3: Presence of ANY of the following: hyperemia, pericholecystic fluid, adhesions to the body, distended gallbladder. 
4: Presence of ANY of the following: adhesions obscuring majority of gallbladder, Grade 1-3 with abnormal liver anatomy, intrahepatic gallbladder, or impacted stone (Mirrizi). 
5: Presence of ANY of the following: perforation, necrosis, inability to visualize the gallbladder due to adhesions. 
Instructions: Assess the image carefully and classify the severity. Only output the severity in a JSON format, eg: \{`severity': 1\}. \\
    \cmidrule(lr){2-3}
    & PaliGemma & ``Answer en What is the severity of inflammation in the provided image of the gallbladder? Choose one number from this list: 1: Normal appearing gallbladder (``robins egg blue") and no adhesions;
2: Minor adhesions at neck otherwise normal gallbladder;
3: Presence of hyperemia pericholecystic fluid adhesions to the body or distended gallbladder;
4: Presence of adhesions obscuring majority of gallbladder  with abnormal liver anatomy intrahepatic gallbladder or impacted stone (Mirrizi); 
5: Presence of perforation or necrosis inability to visualize the gallbladder due to adhesions." \\
    \cmidrule(lr){2-3}
    & CLIP, OpenCLIP & [``Normal appearing gallbladder (`robins egg blue'), no adhesions present, completely normal gallbladder",
    ``Normal appearing gallbladder with minor adhesions at neck or lower part of the gallbladder",
    ``Gallbladder with hyperemia, gallbladder with hyperemiapericholecystic fluid, gallbladder with adhesions to the body, or distended gallbladder", 
    ``Adhesions obscuring majority of gallbladder, with abnormal liver anatomy, intrahepatic gallbladder, or impacted stone (Mirrizi)",
    ``Perforation, necrosis, inability to visualize the gallbladder due to adhesions"] \\
    \cmidrule(lr){2-3}
    & SurgVLP & 
        [``I see a normal appearing gallbladder (`robins egg blue'), no adhesions present, completely normal gallbladder",
        ``I see a normal appearing gallbladder with minor adhesions at neck or lower part of the gallbladder",
        ``I see a gallbladder with hyperemia, gallbladder with hyperemiapericholecystic fluid, gallbladder with adhesions to the body, or distended gallbladder", 
        ``I see adhesions obscuring majority of gallbladder, with abnormal liver anatomy, intrahepatic gallbladder, or impacted stone (Mirrizi)",
        ``I see perforation, necrosis, inability to visualize the gallbladder due to adhesions"]\\
    \midrule
\multirow{1}{*}{Exposure AL} 
    & GPT, Gemini, InternVL2, Phi-3.5 Vision & ``You are a helpful medical video assistant. You will be provided with separate frames uniformaly sampled from a video. Task: predict the laparoscope motion that will occur immediately after the video. The seven types of defined motion are: Static, Up, Down, Left, Right, Zoom-in, Zoom-out. The future movement will be made to ensure proper field-of-view for the surgeon. If no movement is needed, then output Static. Instructions: assess the video carefully, and respond with the future laparoscope movement. Only output one of the given motions, and do not explain why." \\
    \cmidrule(lr){2-3}
    & Qwen2-VL & ``You are a helpful medical video assistant. Task: predict the laparoscope motion that will occur immediately after the video. The seven types of defined motion are: Static, Up, Down, Left, Right, Zoom-in, Zoom-out. The future movement will be made to ensure proper field-of-view for the surgeon. If no movement is needed, then output Static. Instructions: assess the video carefully, and respond with the future laparoscope movement. Only output one of the given motions, and do not explain why." \\
    \midrule
\multirow{1}{*}{Skill HC} 
    & GPT, Gemini, InternVL2, Phi-3.5 Vision &  ``You are a helpful medical video assistant. You will be provided with separate frames uniformaly sampled from a video. Task: assess the tissue handling of a laparoscopic cholecystectomy. It is scored on a scale from 1 to 5. Use the following criteria to output the score: 1. Rough movements, tears tissue, injures adjacent structures, poor grasper control, grasper frequently slips; 3. Handles tissues reasonably well, minor trauma to adjacent tissue (ie, occasional unnecessary bleeding or slipping of the grasper); 5. Handles tissues well, applies appropriate traction, negligible injury to adjacent structures; Instructions: assess the video carefully, and respond with the respect for tissue score. Only output the score."\\
    \cmidrule(lr){2-3}
    & Qwen2-VL & ``You are a helpful medical video assistant. Task: assess the tissue handling of a laparoscopic cholecystectomy. It is scored on a scale from 1 to 5. Use the following criteria to output the score: 1. Rough movements, tears tissue, injures adjacent structures, poor grasper control, grasper frequently slips; 3. Handles tissues reasonably well, minor trauma to adjacent tissue (ie, occasional unnecessary bleeding or slipping of the grasper); 5. Handles tissues well, applies appropriate traction, negligible injury to adjacent structures; Instructions: assess the video carefully, and respond with the respect for tissue score. Only output the score."\\
    \midrule
\multirow{1}{*}{\makecell[l]{Skill IM\\(Tissue\\Handling)}} & GPT, Gemini, Qwen2-VL, InternVL2, Phi-3.5 Vision & "You are a helpful medical video assistant. \
Task: Assess the skill of Tissue Handling for a surgeon provided frames from a cholecystectomy video. Skill is rated on a scale from 1 to 5. \
The rating scale is as follows: \
    - 1. Does not demonstrate careful tissue injury by always showing inappropriate instrument use or unnecessary force \
    - 2. Sometimes demonstrates careful tissue handling, but still often showing instances of unnecessary force or unintentional tissue injury \
    - 3. Usually demonstrates careful tissue handling with sometimes inadvertent tissue injury \
    - 4. Careful tissue handling with only rarely instances of unnecessary force or unintentional tissue injury \
    - 5. Consistently demonstrates careful tissue handling, no tissue injury or unnecessary force \
Instructions: Assess the video clip carefully, and give a 1-5 rating. Use this JSON schema: {"score": int} and avoid line breaks." \\
\cmidrule(lr){2-3}
\multirow{1}{*}{\makecell[l]{(Psykomotor\\Skills)}} & GPT, Gemini, Qwen2-VL, InternVL2, Phi-3.5 Vision &  "You are a helpful medical video assistant. \
Task: Assess the skill of Psykomotor Skills for a surgeon provided frames from a cholecystectomy video. Skill is rated on a scale from 1 to 5. \
The rating scale is as follows: \
    - 1. Struggles with movement coordination and overshooting of instruments for most of step with tentative awkward movements \
    - 2. Some proficiency with use of instruments but often struggles with movement coordination \
    - 3. Developing proficiency in use of both instruments sometimes struggles with movement coordination \
    - 4. Smooth use of instruments rarely with minor deviations in movement coordination and visualization of instrument tips \
    - 5. Fluid movements with both instruments always keeping tips in view \
Instructions: Assess the video clip carefully, and give a 1-5 rating. Use this JSON schema: {"score": int} and avoid line breaks."\\
\cmidrule(lr){2-3}
\multirow{1}{*}{\makecell[l]{(Efficiency)}} & GPT, Gemini, Qwen2-VL, InternVL2, Phi-3.5 Vision &  "You are a helpful medical video assistant. \
Task: Assess the skill of Efficiency for a surgeon provided frames from a cholecystectomy video. Skill is rated on a scale from 1 to 5. \
The rating scale is as follows: \
    - 1. Often encounters disruptions in the flow of progress, frequently pausing and showing uncertainty about the next moves, leading to unnecessary actions and delays \
    - 2. Experiences interruptions in forward progression, sometimes halting operations and displaying uncertainty about the next tasks \
    - 3. Incorporates some forward planning and maintains a reasonably structured progression of the task \
    - 4. Displays proactive planning and anticipates upcoming tasks, resulting in an obvious and well-structured approach for the majority of the task \
    - 5. Demonstrates a clear and well-thought-out plan of action throughout the whole step, foreseeing and seamlessly transitioning to the next task in a planned manner \
Instructions: Assess the video clip carefully, and give a 1-5 rating. Use this JSON schema: {"score": int} and avoid line breaks."\\
\cmidrule(lr){2-3}
\multirow{1}{*}{\makecell[l]{(Dissection\\Quality)}} & GPT, Gemini, Qwen2-VL, InternVL2, Phi-3.5 Vision &  "You are a helpful medical video assistant. \
Task: Assess the skill of Dissection Quality for a surgeon provided frames from a cholecystectomy video. Skill is rated on a scale from 1 to 5. \
The rating scale is as follows: \
    - 1. Consistently operates in the wrong tissue plane, with inadequate correction, resulting in unintended bile spillage and/or bleeding \
    - 2. Struggles with maintaining the tissue plane, often losing it and requiring time to correct, resulting in bile spillage and/or bleeding \
    - 3. Sometimes  experiences a loss of the tissue plane but quickly corrects it, there may be minimal bile spillage and/or bleeding \
    - 4. Rarely deviates into the wrong tissue plane and promptly corrects if any deviation occurs \
    - 5. Demonstrates exceptional consistency in maintaining the correct tissue plane throughout the procedure \
Instructions: Assess the video clip carefully, and give a 1-5 rating. Use this JSON schema: {"score": int} and avoid line breaks."\\
\cmidrule(lr){2-3}
\multirow{1}{*}{\makecell[l]{(Exposure\\Quality)}} & GPT, Gemini, Qwen2-VL, InternVL2, Phi-3.5 Vision &  "You are a helpful medical video assistant. \
Task: Assess the skill of Exposure Quality for a surgeon provided frames from a cholecystectomy video. Skill is rated on a scale from 1 to 5. \
The rating scale is as follows: \
    - 1. Fails to demonstrate landmarks. Poor views and traction. Closed tissue planes and no retraction \
    - 2. Ineffective demonstration of landmarks. Traction often in a suboptimal angle direction. Only little tension on the tissue \
    - 3. Usually demonstrates most landmarks; sometimes with optimal traction and tension on tissue \
    - 4. Demonstrates most landmarks with optimial traction and tension often on tissue \
    - 5. Clearly demonstrates all landmarks. Always with optimal traction and tension throughout \
Instructions: Assess the video clip carefully, and give a 1-5 rating. Use this JSON schema: {"score": int} and avoid line breaks."\\
    \midrule
\multirow{1}{*}{\makecell[l]{Knot Ski JS \\Needle Ski JS\\Suture Ski JS}}
    & GPT, Gemini, InternVL2, Phi-3.5 Vision & ``You are a helpful medical video assistant. You will be provided with separate frames uniformaly sampled from a video. Task: assess the tissue handling of a laparoscopic cholecystectomy. It is scored on a scale from 1 to 5. Use the following criteria to output the score: 1. Rough movements, tears tissue, injures adjacent structures, poor grasper control, grasper frequently slips; 3. Handles tissues reasonably well, minor trauma to adjacent tissue (ie, occasional unnecessary bleeding or slipping of the grasper); 5. Handles tissues well, applies appropriate traction, negligible injury to adjacent structures; Instructions: assess the video carefully, and respond with the respect for tissue score. Only output the score." \\
    \cmidrule(lr){2-3}
    & Qwen2-VL & ``You are a helpful medical video assistant. Task: assess the tissue handling of a laparoscopic cholecystectomy. It is scored on a scale from 1 to 5. Use the following criteria to output the score: 1. Rough movements, tears tissue, injures adjacent structures, poor grasper control, grasper frequently slips; 3. Handles tissues reasonably well, minor trauma to adjacent tissue (ie, occasional unnecessary bleeding or slipping of the grasper); 5. Handles tissues well, applies appropriate traction, negligible injury to adjacent structures; Instructions: assess the video carefully, and respond with the respect for tissue score. Only output the score." \\
    \midrule
\multirow{1}{*}{\makecell[l]{E-Clf IM \\ E-Clf HC \\ E-Clf SST \\ E-Clf C80}} 
    & GPT, Gemini & ``You are a helpful medical video assistant. Task: Classify which type of error occurs in the provided frames from a cholecystectomy video. The errors include: - 1. Bleeding is defined as blood flowing\/moving from a source of injury that is clearly visible on the screen.- 2. Bile spillage is defined as bile spilling out of the gallbladder or biliary ducts. - 3. Thermal injury is defined as an unintentional burn that leads to injury of non-target tissue. - 4. Perforation is defined any tool tissue interaction that leads to perforation of the gallbladder or biliary ducts and the spillage of bile. Use this JSON schema: \{`error\textunderscore{}type': int\} with the type of error (1 for Bleeding, 2 for Bile Spillage, 3 for Thermal Injury, 4 for Perforation) and avoid line breaks. Only return this JSON." \\
    \cmidrule(lr){2-3}
    & Phi-3.5 Vision, InternVL2, Qwen2-VL & ``You are a helpful medical video assistant. You will be provided with separate frames uniformaly sampled from a video segment. Task: classify the surgical error in the video segment. Below are the defined errors: 1. Bleeding 2. Bile spillage 3. Thermal injury 4. Perforation Instructions: Assess the images carefully and classify the error. The segment only contains one error. Only output the error in a JSON format, eg: \{`error\textunderscore{}type': 1\}." \\
    \midrule

\multirow{1}{*}{\makecell[l]{E-Det IM \\ E-Det HC \\ E-Det SST \\ E-Det C80}} 
    & GPT, Gemini & ``You are a helpful medical video assistant. Task: Detect when $<$ERROR\textunderscore{}TYPE$>$ occurs in the provided frames from a cholecystectomy video. Bleeding is defined as blood flowing or moving from the source of injury that is clearly visible on the screen.Bile spillage is defined as containing the first tool tissue interaction that leads to perforation of the gallbladder or biliary ducts and the spillage of bile. Instructions: Assess this 3 minute video clip carefully, and give timestamps (MM:SS) of when the error begins and ends. Assume that there is only one error instance in the video (and there must be one), and the video is recorded at 10 fps. Note that the error can occur for any duration within the video (even the entire 3 minute video). Use this JSON schema: \{`start\textunderscore{}time': MM:SS, `end\textunderscore{}time': MM:SS\} and avoid line breaks. Make sure to give precise timestamps. Only return this JSON." \textit{ERROR\textunderscore{}TYPE}$\in$\{\textit{bleeding, bile spillage}\}\\
    \cmidrule(lr){2-3}
    & Phi-3.5 Vision, InternVL2 & ``You are a helpful medical video assistant. Task: Detect when $<$ERROR\textunderscore{}TYPE$>$ occurs in the provided frames from a cholecystectomy video. Bleeding is defined as blood flowing or moving from the source of injury that is clearly visible on the screen.Bile spillage is defined as containing the first tool tissue interaction that leads to perforation of the gallbladder or biliary ducts and the spillage of bile. Instructions: Assess this video, which is 32 frames sampled from a 3 minute video, and give timestamps (MM:SS) of when the error begins and ends in the original video. Assume that there is only one error instance in the video (and there must be one). Note that the error can occur for any duration within the video (even the entire 3 minute video). Use this JSON schema: \{`start\textunderscore{}time': MM:SS, `end\textunderscore{}time': MM:SS\} and avoid line breaks. Make sure to give precise timestamps. Only return this JSON." \textit{ERROR\textunderscore{}TYPE}$\in$\{\textit{bleeding, bile spillage}\}\\
\cmidrule(lr){2-3}
& Qwen2-VL  & ``This video is 3 minutes long.
Each frame is associated with a specific timestamp using the format `mm:ss'.
Here are the frames and their timestamps:
Frame 0: 00:00
Frame 1: 00:51
Frame 2: 01:42
...
Frame 35: 03:00 (max timestamp)
Given the query: $<$ERROR\textunderscore{}TYPE$>$, when does the described content occur in the video?
Use the `mm:ss' format for your answer. Return in JSON format: {``start": mm:ss, ``end": mm:ss}. 
Only return this JSON." \textit{ERROR\textunderscore{}TYPE}$\in$\{\textit{bleeding, bile spillage}\}\\

\end{longtable}}

\clearpage 
\section{Few-shot Prompts}
\label{sec:prompts}
For few-shot prompting we sample examples, such that each class is see at least X-times.  As some images have multiple labels (for instance the action could be both grasp and cut in one image), one image can cover one instance of several classes. Therefore the number of images is not necessarily X times the number of classes.
\renewcommand{\arraystretch}{1.3}  
{\footnotesize
\begin{longtable}[h!]{p{1.5cm} p{1.8cm} p{11cm}}
    \caption{List of prompts for few-shot experiments. }
    \label{tab:few_prompts}\\
    \toprule
    Pair & Models & Prompt \\
    \midrule
    \endfirsthead

    \toprule
    Pair & Models & Prompt \\
    \midrule
    \endhead

    \midrule
    \multicolumn{3}{r}{\textit{Continued on next page...}} \\
    \midrule
    \endfoot

    \bottomrule
    \endlastfoot

\multirow{1}{*}{\makecell[l]{Tool C80\\ 1-shot}}
    & GPT, Gemini
    & [$<$im$>$, 'output: \{"Grasper": bool, "Bipolar": bool, "Hook": bool, "Scissors": bool, "Clipper": bool, "Irrigator": bool, "SpecimenBag": bool\}', ...,  
 'You just saw some images of a laparoscopic cholecystectomy with corresponding tool annotations. In the next image, identify which of these tools is present in an image: Grasper, Bipolar, Hook, Scissors, Clipper, Irrigator, SpecimenBag? Respond with a 0 or 1 for all tools according to whether or not the tool is present. Use this JSON schema: \{``tool\textunderscore{}name": bool"\} and avoid line breaks.'$]$ \textit{All .jpg images are loaded before passed to API}. \textit{List of images}: 
/train/video01/8900.jpg,
/train/video05/32250.jpg,
/train/video01/675.jpg,
/train/video01/9525.jpg,
/train/video07/106850.jpg,
/train/video03/111875.jpg,
/train/video06/8200.jpg.
 \\
    \cmidrule(lr){2-3}
    \multirow{1}{*}{\makecell[l]{Tool C80\\ 3-shot}}
    & GPT, Gemini
    & \textit{Prompt above with this list of images}: 
/train/video37/9625.jpg, /train/video11/18725.jpg, /train/video16/37425.jpg, /train/video10/18650.jpg, /train/video07/110275.jpg, /train/video14/40600.jpg, /train/video33/31750.jpg, /train/video26/19650.jpg, /train/video18/43400.jpg, /train/video35/33600.jpg, /train/video18/45350.jpg, /train/video40/36050.jpg, /train/video24/35125.jpg, /train/video03/114025.jpg, /train/video08/19450.jpg, /train/video03/113850.jpg.
 \\
    \cmidrule(lr){2-3}
    \multirow{1}{*}{\makecell[l]{Tool C80\\ 5-shot}}
    & GPT, Gemini
    & \textit{Prompt above with this list of images}: /train/video13/925.jpg, /train/video04/26650.jpg, /train/video26/10575.jpg, /train/video26/19925.jpg, /train/video05/29800.jpg, /train/video06/21475.jpg, /train/video01/5175.jpg, /train/video26/22100.jpg, /train/video38/12525.jpg, /train/video07/54600.jpg, /train/video14/33300.jpg, /train/video40/36200.jpg, /train/video17/27925.jpg, /train/video02/40975.jpg, /train/video18/45100.jpg, /train/video14/29825.jpg, /train/video29/16200.jpg, /train/video39/22175.jpg, /train/video06/33550.jpg, /train/video31/95825.jpg, /train/video18/42000.jpg, /train/video06/34375.jpg, /train/video32/3725.jpg, /train/video27/46925.jpg, /train/video11/2750.jpg, /train/video24/47700.jpg, /train/video30/22200.jpg, /train/video36/58450.jpg, /train/video34/31275.jpg, /train/video18/25125.jpg, /train/video03/111975.jpg, /train/video23/17475.jpg.

 \\
    \midrule
  \multirow{1}{*}{\makecell[l]{Tool HC\\ 1-shot}}
    & GPT, Gemini
    & [$<$im$>$, 'output: \{"Grasper": bool, "Clipper": bool, "Coagulation instruments": bool, "Scissors": bool, "Suction-irrigation": bool, "Specimen bag": bool, "Stapler": bool\}', ...,  
 'You just saw some images of a laparoscopic cholecystectomy with corresponding tool annotations. In the next image, assess which of these tools is present: Grasper, Clipper, Coagulation instruments, Scissors, Suction-irrigation, Specimen bag, Stapler.         Respond with a 0 or 1 for all tools according to whether or not the tool is present.         Use this JSON schema: {`tool\textunderscore{}name': bool} and avoid line breaks.'$]$ \textit{All .png images are loaded before passed to API}. List of images: 
Hei-Chole12/frame\textunderscore{}09234.png, Hei-Chole11/frame\textunderscore{}08049.png, Hei-Chole11/frame\textunderscore{}10931.png, Hei-Chole11/frame\textunderscore{}21704.png, Hei-Chole11/frame\textunderscore{}31351.png, Hei-Chole24/frame\textunderscore{}95118.png, Hei-Chole12/frame\textunderscore{}00003.png.
 \\
    \cmidrule(lr){2-3} 
    \multirow{1}{*}{\makecell[l]{Tool HC\\ 3-shot}}
    & GPT, Gemini
    & \textit{Prompt above with this list of images}: 
Hei-Chole24/frame\textunderscore{}74625.png, Hei-Chole24/frame\textunderscore{}12375.png, Hei-Chole24/frame\textunderscore{}11625.png, Hei-Chole12/frame\textunderscore{}12000.png, Hei-Chole12/frame\textunderscore{}40500.png, Hei-Chole24/frame\textunderscore{}48375.png, Hei-Chole24/frame\textunderscore{}109125.png, Hei-Chole12/frame\textunderscore{}42750.png, Hei-Chole24/frame\textunderscore{}96000.png, Hei-Chole12/frame\textunderscore{}43125.png, Hei-Chole24/frame\textunderscore{}100500.png, Hei-Chole24/frame\textunderscore{}102375.png, Hei-Chole12/frame\textunderscore{}09375.png, Hei-Chole24/frame\textunderscore{}95625.png, Hei-Chole11/frame\textunderscore{}14625.png, Hei-Chole24/frame\textunderscore{}67875.png, Hei-Chole24/frame\textunderscore{}97500.png, Hei-Chole12/frame\textunderscore{}10500.png, Hei-Chole24/frame\textunderscore{}93000.png, Hei-Chole12/frame\textunderscore{}25875.png, Hei-Chole24/frame\textunderscore{}74250.png
 \\
    \cmidrule(lr){2-3}
    \multirow{1}{*}{\makecell[l]{Tool HC\\ 5-shot}}
    & GPT, Gemini
    & \textit{Prompt above with this list of images}: Hei-Chole24/frame\textunderscore{}48750.png, Hei-Chole11/frame\textunderscore{}07125.png, Hei-Chole24/frame\textunderscore{}45000.png, Hei-Chole24/frame\textunderscore{}77625.png, Hei-Chole12/frame\textunderscore{}37125.png, Hei-Chole24/frame\textunderscore{}76500.png, Hei-Chole24/frame\textunderscore{}49500.png, Hei-Chole12/frame\textunderscore{}42375.png, Hei-Chole12/frame\textunderscore{}35250.png, Hei-Chole11/frame\textunderscore{}18750.png, Hei-Chole24/frame\textunderscore{}110250.png, Hei-Chole24/frame\textunderscore{}46125.png, Hei-Chole11/frame\textunderscore{}09375.png, Hei-Chole12/frame\textunderscore{}40875.png, Hei-Chole24/frame\textunderscore{}102375.png, Hei-Chole24/frame\textunderscore{}107625.png, Hei-Chole24/frame\textunderscore{}124500.png, Hei-Chole11/frame\textunderscore{}14625.png, Hei-Chole12/frame\textunderscore{}41625.png, Hei-Chole12/frame\textunderscore{}22500.png, Hei-Chole24/frame\textunderscore{}67875.png, Hei-Chole12/frame\textunderscore{}09000.png, Hei-Chole24/frame\textunderscore{}92625.png, Hei-Chole24/frame\textunderscore{}96000.png, Hei-Chole11/frame\textunderscore{}10875.png, Hei-Chole24/frame\textunderscore{}96750.png, Hei-Chole12/frame\textunderscore{}25875.png, Hei-Chole12/frame\textunderscore{}10125.png, Hei-Chole24/frame\textunderscore{}93000.png, Hei-Chole24/frame\textunderscore{}95250.png, Hei-Chole24/frame\textunderscore{}92250.png, Hei-Chole24/frame\textunderscore{}74250.png, Hei-Chole24/frame\textunderscore{}95625.png, Hei-Chole12/frame\textunderscore{}10875.png, Hei-Chole24/frame\textunderscore{}97500.png.
 \\
    \midrule
  \multirow{1}{*}{\makecell[l]{Tool ES\\ Anat ES \\ 1-shot}}
    & GPT, Gemini
    & $[$'train/8\textunderscore{}14775.jpg', 'output: \{"calot triangle": [433, 530, 547, 675], "cystic artery": [0, 590, 500, 740], "cystic duct": [466, 422, 779, 720], "gallbladder": [0, 81, 645, 740], "tool1": [54, 685, 420, 998], "tool2": [187, 2, 439, 222]\}', 'You just saw bounding boxes for anatomies and tools in an image. For the next image, return bounding boxes for the cystic artery, cystic duct, cystic plate, gallbladder and all tools [ymin, xmin, ymax, xmax] if they are present.'$]$ \textit{All .jpg images are loaded before passed to API.}\\
    \cmidrule(lr){2-3} 
  \multirow{1}{*}{\makecell[l]{Tool ES\\ Anat ES \\ 3-shot}}
    & GPT, Gemini
    & ['train/8\textunderscore{}14775.jpg', 'output: \{"calot triangle": [433, 530, 547, 675], "cystic artery": [0, 590, 500, 740], "cystic duct": [466, 422, 779, 720], "gallbladder": [0, 81, 645, 740], "tool1": [54, 685, 420, 998], "tool2": [187, 2, 439, 222]\}', 'train/11\textunderscore{}26275.jpg', 'output: \{"cystic\textunderscore{}plate": [481, 358, 618, 423], "calot\textunderscore{}triangle": [454, 380, 614, 422], "cystic\textunderscore{}artery": [310, 368, 622, 507], "cystic\textunderscore{}duct": [504, 252, 808, 483], "gallbladder": [0, 133, 633, 453], "tool1": [0, 392, 481, 788], "tool2": [452, 132, 575, 270]\}', 'train/8\textunderscore{}16275.jpg', 'output: \{"cystic\textunderscore{}artery": [0, 510, 275, 614], "cystic\textunderscore{}duct": [302, 357, 616, 573], "gallbladder": [0, 134, 450, 512], "tool": [16, 1, 210, 329]\}', 'You just saw bounding boxes for anatomies and tools in an image. For the next image, return bounding boxes for the cystic artery, cystic duct, cystic plate, gallbladder and all tools [ymin, xmin, ymax, xmax] if they are present.'] \textit{All .jpg images are loaded before passed to API.}\\
    \cmidrule(lr){2-3} 
  \multirow{1}{*}{\makecell[l]{Tool ES\\ Anat ES \\5-shot}}
    & GPT, Gemini
    & ['train/8\textunderscore{}14775.jpg', 'output: \{"calot triangle": [433, 530, 547, 675], "cystic artery": [0, 590, 500, 740], "cystic duct": [466, 422, 779, 720], "gallbladder": [0, 81, 645, 740], "tool1": [54, 685, 420, 998], "tool2": [187, 2, 439, 222]\}', 'train/11\textunderscore{}26275.jpg', 'output: \{"cystic\textunderscore{}plate": [481, 358, 618, 423], "calot\textunderscore{}triangle": [454, 380, 614, 422], "cystic\textunderscore{}artery": [310, 368, 622, 507], "cystic\textunderscore{}duct": [504, 252, 808, 483], "gallbladder": [0, 133, 633, 453], "tool1": [0, 392, 481, 788], "tool2": [452, 132, 575, 270]\}', 'train/8\textunderscore{}16275.jpg', 'output: \{"cystic\textunderscore{}artery": [0, 510, 275, 614], "cystic\textunderscore{}duct": [302, 357, 616, 573], "gallbladder": [0, 134, 450, 512], "tool": [16, 1, 210, 329]\}', 'train/8\textunderscore{}17025.jpg', 'output: \{"cystic\textunderscore{}artery": [0, 407, 400, 475], "cystic\textunderscore{}duct": [435, 168, 681, 437], "gallbladder": [2, 2, 437, 422], "tool1": [397, 2, 591, 278], "tool2": [445, 481, 864, 998]\}', 'train/8\textunderscore{}17775.jpg', 'output: \{"cystic\textunderscore{}artery": [0, 566, 118, 613], "cystic\textunderscore{}duct": [389, 403, 600, 600], "gallbladder": [0, 215, 379, 581], "tool1": [0, 1, 212, 365], "tool2": [0, 289, 464, 998]\}', 'You just saw bounding boxes for anatomies and tools in an image. For the next image, return bounding boxes for the cystic artery, cystic duct, cystic plate, gallbladder and all tools [ymin, xmin, ymax, xmax] if they are present.'] \textit{All .jpg images are loaded before passed to API.}\\
    \midrule
  \multirow{1}{*}{\makecell[l]{Action AV\\ 1-shot}}
    & GPT, Gemini
    & [$<$im$>$, ``output: \{`action': \textit{int}\}", ..., ``You just saw some images and their coressponding action annotations.  For the next image, determine the action being performed in the image. The possible actions are                                                                 0: cutting - if the surgeon is using a tool like scissors, scalpel, knife, or electrocautery device to cut or dissect tissues.                                                                     1: tying - if the surgeon is using their hands or needle holders to create secure knots.                                                                         2: suturing - if the surgeon is closing an open wound with a needle but not creating secure knots.                                                                             3: background - if no surgical action is being performed, including actions like using forceps, clamps, retractors, or dilators.                                                                                 There are no other options. Use this JSON schema: {`action': int} and avoid line breaks. "] \textit{All .jpg images are loaded before passed to API}. \textit{List of images}: EUdac6A9n60-000006767.jpg, Xg6vD3vngLQ-000000975.jpg, FUGhWj5iv70-000008279.jpg, JUTyS7ZRRkQ-000006353.jpg.
 \\
        \cmidrule(lr){2-3}
    \multirow{1}{*}{\makecell[l]{Action AV\\ 3-shot}}
    & GPT, Gemini
    & \textit{Prompt above with this list of images}: S4p9MmTfVTs-000001445.jpg, RWHBTwfa5C8-000003575.jpg, toE\textunderscore{}4MtsqQM-000000488.jpg, dPvRrcSsc6Y-000000662.jpg, cpgMJ7KOVl8-000004213.jpg, N32N6VEcW2I-000007967.jpg, vtK1XN4ZaU4-000003552.jpg, TFwFMav\textunderscore{}cpE-000014219.jpg, N32N6VEcW2I-000007303.jpg, VtJtGtC3R80-000003911.jpg, SNsUtI82de8-000009903.jpg, e12tIDPDfwU-000003095.jpg.\\

    \cmidrule(lr){2-3}
    \multirow{1}{*}{\makecell[l]{Action AV\\ 5-shot}}
    & GPT, Gemini
    & \textit{Prompt above with this list of images}: EswP8VDC85s-000000799.jpg, synW6molzgA-000005228.jpg, l5h\textunderscore{}tOU\textunderscore{}D9w-000000254.jpg, GJ5RwKonnms-000001211.jpg, L8k75Onag\textunderscore{}o-000006575.jpg, oD5gC2ESBnk-000006759.jpg, FotC4hB7Y0c-000002397.jpg, ou4iO5ah9ys-000000995.jpg, ytgWAMS1SkE-000007629.jpg, V7vkRKaUkn8-000008998.jpg, DpeAsOXVruw-000003135.jpg, 6idNh90AdtA-000001685.jpg, VtJtGtC3R80-000007171.jpg, S1R95eOuSNk-000002039.jpg, S1R95eOuSNk-000004079.jpg, S4p9MmTfVTs-000004337.jpg, 3Ql0fGVrQeA-000003717.jpg, ytgWAMS1SkE-000009809.jpg, LgmXCOICHLA-000002384.jpg, GwHruH8trhg-000001674.jpg.\\
\midrule
      \multirow{1}{*}{\makecell[l]{Action HC\\ 1-shot}}
    & GPT, Gemini
    & [$<$im$>$, `output: \{"grasp": bool, "hold": bool, "cut": bool, "clip": bool\}', ,..., 'You just saw some images of a laparoscopic cholecystectomy with corresponding action annotations. In the next image, find all tools. For each tool, decide which action it performs:     grasp, hold, cut, or clip. It is possible that the insturment is idle and no action is performed. Aggregate the actions across all instruments. For each action return a boolean indicating if the action is performed by any instrument.     Use this JSON schema: {"grasp": bool, "hold": bool, "cut": bool, "clip": bool} and avoid line breaks. ']\textit{All .png images are loaded before passed to API}. \textit{List of images}: Hei-Chole12/frame\textunderscore{}09234.png, Hei-Chole11/frame\textunderscore{}10931.png, Hei-Chole12/frame\textunderscore{}02987.png, Hei-Chole11/frame\textunderscore{}14298.png.\\

        \cmidrule(lr){2-3}
    \multirow{1}{*}{\makecell[l]{Action HC\\ 3-shot}}
    & GPT, Gemini
    & \textit{Prompt above with this list of images}: Hei-Chole24/frame\textunderscore{}112875.png, Hei-Chole24/frame\textunderscore{}79500.png, Hei-Chole24/frame\textunderscore{}84750.png, Hei-Chole24/frame\textunderscore{}117375.png, Hei-Chole24/frame\textunderscore{}109500.png, Hei-Chole11/frame\textunderscore{}13875.png, Hei-Chole11/frame\textunderscore{}26625.png, Hei-Chole12/frame\textunderscore{}44250.png, Hei-Chole11/frame\textunderscore{}14250.png.\\

        \cmidrule(lr){2-3}
    \multirow{1}{*}{\makecell[l]{Action HC\\ 5-shot}}
    & GPT, Gemini
    & \textit{Prompt above with this list of images}: Hei-Chole24/frame\textunderscore{}43125.png, Hei-Chole24/frame\textunderscore{}43875.png, Hei-Chole24/frame\textunderscore{}20250.png, Hei-Chole24/frame\textunderscore{}109875.png, Hei-Chole24/frame\textunderscore{}123750.png, Hei-Chole12/frame\textunderscore{}44250.png, Hei-Chole24/frame\textunderscore{}117375.png, Hei-Chole11/frame\textunderscore{}13875.png, Hei-Chole12/frame\textunderscore{}03375.png, Hei-Chole11/frame\textunderscore{}26625.png, Hei-Chole24/frame\textunderscore{}34875.png, Hei-Chole11/frame\textunderscore{}10125.png, Hei-Chole24/frame\textunderscore{}09375.png, Hei-Chole11/frame\textunderscore{}14250.png.\\
\midrule
  \multirow{1}{*}{\makecell[l]{CVS ES\\ 1-shot}}
    & GPT, Gemini
    & [$<$im$>$, 'output: \{"C1": bool, "C2": bool, "C3": bool\}', ...,  'You just saw some images of a laparoscopic cholecystectomy with corresponding Critical View of Safety annotations. In the next image, assess whether Critical View of Safety (CVS) is fully achieved in the provided frames from a cholecystectomy video.                                                                 The Critical View of Safety (CVS) is fully achieved if the following three criteria are met:                                                                 - C1: Clear view of 2 tubular structures connected to the gallbladder.                                                                 - C2: A carefully dissected hepatocystic triangle presenting an unimpeded view of only the 2 cystic structures and the cystic plate.                                                                 - C3: The lower third of the gallbladder is dissected off the cystic plate.                                                             Instructions: Assess the image carefully, and answer which of the Critical View of Safety (CVS) criteria are met.                                                             Use this JSON schema: {"C1": bool, "C2": bool, "C3": bool} and avoid line breaks.']   \textit{All .jpg images are loaded before passed to API}. \textit{List of images}: train/10\textunderscore{}17850.jpg, train/22\textunderscore{}41275.jpg, train/23\textunderscore{}28600.jpg.\\
            \cmidrule(lr){2-3}
    \multirow{1}{*}{\makecell[l]{CVS ES\\ 3-shot}}
    & GPT, Gemini
    & \textit{Prompt above with this list of images}: train/10\textunderscore{}17850.jpg, train/22\textunderscore{}41275.jpg, train/23\textunderscore{}28600.jpg, train/4\textunderscore{}36950.jpg, train/1\textunderscore{}29850.jpg, train/15\textunderscore{}30775.jpg, train/89\textunderscore{}26875.jpg, train/106\textunderscore{}53450.jpg, train/57\textunderscore{}27650.jpg.\\

            \cmidrule(lr){2-3}
    \multirow{1}{*}{\makecell[l]{CVS ES\\ 5-shot}}
    & GPT, Gemini
    & \textit{Prompt above with this list of images}: train/95\textunderscore{}28500.jpg, train/42\textunderscore{}46900.jpg, train/82\textunderscore{}24600.jpg, train/88\textunderscore{}33075.jpg, train/6\textunderscore{}16425.jpg, train/117\textunderscore{}21625.jpg, train/76\textunderscore{}57225.jpg, train/31\textunderscore{}49400.jpg, train/57\textunderscore{}34400.jpg, train/41\textunderscore{}600.jpg, train/4\textunderscore{}24325.jpg.\\
\midrule
  \multirow{1}{*}{\makecell[l]{Phase C80\\ 1-shot}}
    & GPT, Gemini
    & [$<$im$>$, 'output: \{"phase": \textit{int}\}', ..., 'You just saw some images of a laparoscopic cholecystectomy with corresponding phase annotations. In the next image,                          determine the surgical phase of the image. The possible phases are                          0: Preparation, 1: Calot Triangle Dissection, 2: Clipping Cutting, 3: Gallbladder Dissection,                         4: Gallbladder Packaging, 5: Cleaning Coagulation, 6: Gallbladder Retraction. There are no other options.                         Use this JSON schema: {"phase": \textit{int}} and avoid line breaks.'] \textit{All .jpg images are loaded before passed to API}. \textit{List of images}: train/video01/0.jpg, train/video01/7000.jpg, train/video01/21000.jpg, train/video01/27000.jpg, train/video01/39000.jpg, train/video01/40000.jpg, train/video01/43000.jpg.\\
                \cmidrule(lr){2-3}
    \multirow{1}{*}{\makecell[l]{Phase C80\\ 3-shot}}
    & GPT, Gemini
    & \textit{Prompt above with this list of images}: train/video03/7225.jpg, train/video32/14925.jpg, train/video38/20800.jpg, train/video16/60650.jpg, train/video39/40450.jpg, train/video23/35850.jpg, train/video02/35425.jpg, train/video30/44050.jpg, train/video02/43750.jpg, train/video03/141475.jpg, train/video27/26350.jpg, train/video03/127475.jpg, train/video02/38150.jpg, train/video23/30250.jpg, train/video40/200.jpg, train/video08/30500.jpg, train/video30/25.jpg, train/video30/72250.jpg, train/video08/35675.jpg, train/video32/46450.jpg, train/video15/1650.jpg.\\
                \cmidrule(lr){2-3}
    \multirow{1}{*}{\makecell[l]{Phase C80\\ 5-shot}}
    & GPT, Gemini
    & \textit{Prompt above with this list of images}: train/video07/83500.jpg, train/video12/14100.jpg, train/video18/19525.jpg, train/video30/15600.jpg, train/video17/14125.jpg, train/video02/2775.jpg, train/video03/134600.jpg, train/video30/34850.jpg, train/video35/41075.jpg, train/video07/107675.jpg, train/video34/9475.jpg, train/video07/83350.jpg, train/video30/43900.jpg, train/video03/130575.jpg, train/video35/3725.jpg, train/video17/1975.jpg, train/video09/62575.jpg, train/video30/21900.jpg, train/video02/33400.jpg, train/video05/31200.jpg, train/video18/19975.jpg, train/video04/3225.jpg, train/video32/48025.jpg, train/video20/29450.jpg, train/video35/1125.jpg, train/video29/44050.jpg, train/video38/62400.jpg, train/video24/6450.jpg, train/video20/28100.jpg, train/video12/22950.jpg, train/video27/46575.jpg, train/video06/53050.jpg, train/video05/53650.jpg, train/video22/36600.jpg, train/video34/32725.jpg.\\
    \midrule
      \multirow{1}{*}{\makecell[l]{Phase HC\\ 1-shot}}
    & GPT, Gemini
    & [$<$im$>$, 'output: \{"phase": int\}',  'You just saw some images of a laparoscopic cholecystectomy and their coressponding phase annotations.  For the next image, determine the surgical phase of the image. The possible phases are     0: Preparation, 1: Calot Triangle Dissection, 2: Clipping Cutting, 3: Gallbladder Dissection,     4: Gallbladder Packaging, 5: Cleaning Coagulation, 6: Gallbladder Retraction. There are no other options. Use this JSON schema: \{"phase": int\} and avoid line breaks.'] \textit{All .png images are loaded before passed to API}. \textit{List of images}: Hei-Chole11/frame\textunderscore{}00000.png, Hei-Chole12/frame\textunderscore{}12569.png, Hei-Chole12/frame\textunderscore{}09762.png, Hei-Chole11/frame\textunderscore{}20595.png, Hei-Chole12/frame\textunderscore{}12569.png, Hei-Chole24/frame\textunderscore{}116426.png, Hei-Chole24/frame\textunderscore{}123686.png.\\
                \cmidrule(lr){2-3}
    \multirow{1}{*}{\makecell[l]{Phase HC\\ 3-shot}}
    & GPT, Gemini
    & \textit{Prompt above with this list of images}: Hei-Chole11/frame\textunderscore{}18950.png, Hei-Chole24/frame\textunderscore{}121100.png, Hei-Chole12/frame\textunderscore{}25575.png, Hei-Chole24/frame\textunderscore{}36700.png, Hei-Chole24/frame\textunderscore{}40025.png, Hei-Chole24/frame\textunderscore{}124450.png, Hei-Chole24/frame\textunderscore{}50325.png, Hei-Chole24/frame\textunderscore{}00900.png, Hei-Chole11/frame\textunderscore{}25125.png, Hei-Chole24/frame\textunderscore{}128050.png, Hei-Chole11/frame\textunderscore{}34925.png, Hei-Chole24/frame\textunderscore{}89050.png, Hei-Chole11/frame\textunderscore{}12500.png, Hei-Chole12/frame\textunderscore{}01550.png, Hei-Chole12/frame\textunderscore{}34550.png, Hei-Chole24/frame\textunderscore{}123600.png, Hei-Chole24/frame\textunderscore{}110625.png, Hei-Chole24/frame\textunderscore{}03925.png, Hei-Chole24/frame\textunderscore{}93400.png, Hei-Chole24/frame\textunderscore{}100025.png, Hei-Chole24/frame\textunderscore{}105775.png.\\
                    \cmidrule(lr){2-3}
    \multirow{1}{*}{\makecell[l]{Phase HC\\ 5-shot}}
    & GPT, Gemini
    & \textit{Prompt above with this list of images}:Hei-Chole24/frame\textunderscore{}124700.png, Hei-Chole24/frame\textunderscore{}22650.png, Hei-Chole24/frame\textunderscore{}106975.png, Hei-Chole24/frame\textunderscore{}33650.png, Hei-Chole24/frame\textunderscore{}106150.png, Hei-Chole11/frame\textunderscore{}01275.png, Hei-Chole24/frame\textunderscore{}39275.png, Hei-Chole12/frame\textunderscore{}02825.png, Hei-Chole24/frame\textunderscore{}110675.png, Hei-Chole24/frame\textunderscore{}20275.png, Hei-Chole12/frame\textunderscore{}30325.png, Hei-Chole24/frame\textunderscore{}116700.png, Hei-Chole24/frame\textunderscore{}118050.png, Hei-Chole24/frame\textunderscore{}108100.png, Hei-Chole12/frame\textunderscore{}24300.png, Hei-Chole24/frame\textunderscore{}91300.png, Hei-Chole24/frame\textunderscore{}109075.png, Hei-Chole24/frame\textunderscore{}02575.png, Hei-Chole24/frame\textunderscore{}103275.png, Hei-Chole24/frame\textunderscore{}94250.png, Hei-Chole12/frame\textunderscore{}29925.png, Hei-Chole11/frame\textunderscore{}35175.png, Hei-Chole12/frame\textunderscore{}43725.png, Hei-Chole12/frame\textunderscore{}30825.png, Hei-Chole24/frame\textunderscore{}100550.png, Hei-Chole12/frame\textunderscore{}24575.png, Hei-Chole24/frame\textunderscore{}124875.png, Hei-Chole12/frame\textunderscore{}37700.png, Hei-Chole11/frame\textunderscore{}14175.png, Hei-Chole11/frame\textunderscore{}02600.png, Hei-Chole24/frame\textunderscore{}125500.png, Hei-Chole24/frame\textunderscore{}122425.png, Hei-Chole24/frame\textunderscore{}126325.png, Hei-Chole24/frame\textunderscore{}05225.png, Hei-Chole24/frame\textunderscore{}03350.png. \\

\end{longtable}

\section{Additional Qualitative Examples}
\begin{figure}[H]
\begin{center}
    \includegraphics[width=0.94\linewidth]{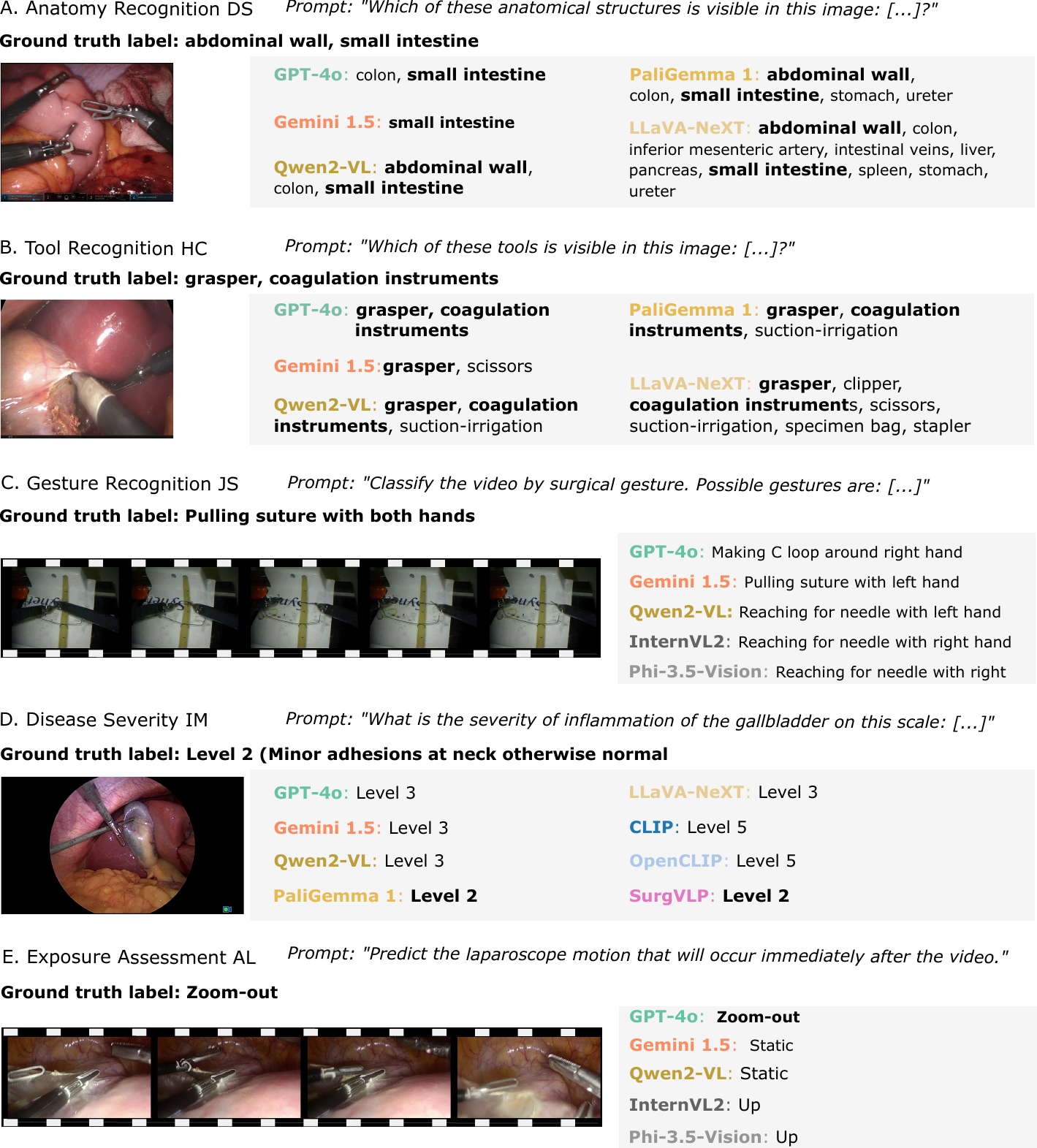}
    \end{center}
        \footnotesize{
    Datasets: DS=Dresden Surgical Anatomy, HC=HeiChole, JS=JIGSAWS, IM=Intermountain, AL=AutoLaparo.}
    \caption{Additional qualitative examples. Qualitative zero-shot results for various tasks, models, and datasets. Correct predictions are shown in \textbf{bold}. Prompts shortened for display; full versions in Section \ref{sec:prompts}. \textbf{A)} General-purpose VLMs successfully identify anatomies; however, models like LLaVA-NeXT tend to over-predict. \textbf{B)} GPT-4o leads at tool recognition. \textbf{C)} Gesture recognition is an unsolved problem. \textbf{D)} SurgVLP leads at disease severity assessment. \textbf{E)} GPT-4o leads at exposure assessment.}
    \label{fig:quali_supp}
\end{figure}

\section{Results with Additional Metrics}
\normalsize

In this section we display the main results from Figure \ref{fig:main} in Tables \ref{tab:PhaseMB} - \ref{tab:E-DetIM}. In addition to the F1 Score (F1), we also report Accuracy (A), Jaccard Score (J), Precision (P), and Recall (R). For all metrics except accuracy, we also report the weighted metric (w) that accounts for class imbalance.

\begin{table*}[h]
\centering
\caption{Phase MB}\label{tab:PhaseMB}\setlength{\tabcolsep}{4pt}
\begin{tabular}{lccccc|cccc}
Model & A & F1 & J & P & R & wF1 & wJ & wP & wR \\
\hline
CLIP & 0.03 & 0.00 & 0.00 & 0.02 & 0.08 & 0.00 & 0.00 & 0.06 & 0.03 \\
OpenCLIP & 0.25 & 0.08 & 0.04 & 0.12 & 0.10 & 0.19 & 0.11 & 0.24 & 0.25 \\
Qwen2-VL & 0.05 & 0.01 & 0.01 & 0.10 & 0.09 & 0.02 & 0.01 & 0.16 & 0.05 \\
SurgVLP & 0.14 & \textbf{0.13} & \textbf{0.07} & 0.15 & 0.16 & 0.13 & 0.08 & 0.20 & 0.14 \\
Gemini-1.5-Pro & 0.17 & 0.06 & 0.04 & 0.07 & 0.10 & 0.15 & 0.09 & 0.19 & 0.17 \\
GPT-4o & \textbf{0.28} & 0.11 & 0.06 & \textbf{0.18} & \textbf{0.18} & \textbf{0.21} & \textbf{0.13} & \textbf{0.30} & \textbf{0.28} \\
LLaVA-NeXT & 0.20 & 0.03 & 0.02 & 0.10 & 0.08 & 0.07 & 0.04 & \textbf{0.30} & 0.20 \\
PaliGemma & 0.05 & 0.02 & 0.01 & 0.01 & 0.09 & 0.01 & 0.01 & 0.01 & 0.05 \\
\end{tabular}
\end{table*}

\begin{table*}[h]
\centering
\caption{E-Clf IM}\label{tab:E-ClfIM}\setlength{\tabcolsep}{4pt}
\begin{tabular}{lccccc|cccc}
Model & A & F1 & J & P & R & wF1 & wJ & wP & wR \\
\hline
InternVL2 & 0.21 & 0.10 & 0.06 & 0.12 & 0.24 & 0.10 & 0.06 & 0.19 & 0.21 \\
Phi-3.5 Vision & 0.21 & 0.14 & 0.08 & 0.09 & 0.29 & 0.10 & 0.06 & 0.06 & 0.21 \\
Qwen2-VL & 0.21 & 0.09 & 0.05 & 0.05 & 0.25 & 0.07 & 0.04 & 0.04 & 0.21 \\
Gemini-1.5-Pro & 0.25 & 0.21 & 0.12 & 0.36 & 0.32 & 0.21 & 0.12 & \textbf{0.57} & 0.25 \\
GPT-4o & \textbf{0.37} & \textbf{0.31} & \textbf{0.19} & \textbf{0.42} & \textbf{0.40} & \textbf{0.35} & \textbf{0.22} & \textbf{0.57} & \textbf{0.37} \\
\end{tabular}
\end{table*}

\begin{table*}[h]
\centering
\caption{Tool C80}\label{tab:ToolC80}\setlength{\tabcolsep}{4pt}
\begin{tabular}{lccccc|cccc}
Model & A & F1 & J & P & R & wF1 & wJ & wP & wR \\
\hline
CLIP & 0.68 & 0.20 & 0.15 & 0.17 & 0.43 & 0.57 & 0.43 & 0.44 & \textbf{0.86} \\
Med-Gemini & 0.64 & 0.28 & 0.19 & 0.29 & 0.34 & \textbf{0.67} & \textbf{0.54} & 0.75 & 0.64 \\
OpenCLIP & 0.76 & 0.22 & 0.15 & 0.17 & 0.34 & 0.54 & 0.39 & 0.43 & 0.73 \\
Qwen2-VL & 0.69 & 0.31 & 0.22 & 0.30 & 0.52 & 0.58 & 0.43 & 0.55 & 0.71 \\
SurgVLP & 0.52 & 0.25 & 0.16 & 0.21 & \textbf{0.73} & 0.52 & 0.37 & 0.50 & 0.64 \\
Gemini-1.5-Pro & 0.72 & 0.22 & 0.14 & 0.36 & 0.38 & 0.30 & 0.21 & 0.66 & 0.42 \\
GPT-4o & \textbf{0.89} & \textbf{0.46} & \textbf{0.34} & 0.55 & 0.46 & \textbf{0.67} & 0.52 & \textbf{0.81} & 0.60 \\
LLaVA-NeXT & 0.59 & 0.23 & 0.15 & 0.22 & 0.55 & 0.49 & 0.34 & 0.54 & 0.51 \\
PaliGemma & 0.09 & 0.31 & 0.22 & \textbf{0.69} & 0.30 & 0.50 & 0.36 & 0.57 & 0.54 \\
\end{tabular}
\end{table*}

\begin{table*}
\centering
\caption{Knot Ski JS}\label{tab:jigsaws_skill_assessment_debug_Knot_Tying}\setlength{\tabcolsep}{4pt}
\begin{tabular}{lccccc|cccc}
Model & A & F1 & J & P & R & wF1 & wJ & wP & wR \\
\hline
InternVL2 & 0.11 & 0.04 & 0.03 & 0.04 & 0.06 & 0.08 & 0.05 & 0.07 & 0.11 \\
Phi-3.5 Vision & 0.27 & 0.16 & 0.10 & \textbf{0.20} & 0.27 & 0.17 & 0.11 & \textbf{0.23} & 0.27 \\
Qwen2-VL & \textbf{0.37} & 0.16 & 0.11 & 0.15 & 0.27 & \textbf{0.22} & \textbf{0.15} & 0.19 & \textbf{0.37} \\
Gemini-1.5-Pro & 0.24 & \textbf{0.17} & \textbf{0.11} & 0.17 & 0.26 & 0.17 & 0.11 & 0.17 & 0.24 \\
GPT-4o & 0.36 & 0.14 & 0.10 & 0.10 & \textbf{0.27} & 0.19 & 0.13 & 0.13 & 0.36 \\
\end{tabular}
\end{table*}

\begin{table*}
\centering
\caption{Verb C45}\label{tab:VerbC45}\setlength{\tabcolsep}{4pt}
\begin{tabular}{lccccc|cccc}
Model & A & F1 & J & P & R & wF1 & wJ & wP & wR \\
\hline
CLIP & 0.15 & 0.21 & 0.15 & 0.15 & \textbf{1.00} & \textbf{0.59} & 0.46 & 0.46 & \textbf{1.00} \\
Med-Gemini & 0.25 & 0.12 & 0.07 & 0.22 & 0.14 & 0.31 & 0.18 & \textbf{0.64} & 0.25 \\
OpenCLIP & 0.62 & 0.19 & 0.14 & 0.18 & 0.50 & \textbf{0.59} & \textbf{0.47} & 0.50 & 0.83 \\
Qwen2-VL & 0.72 & 0.08 & 0.05 & 0.17 & 0.23 & 0.15 & 0.08 & 0.54 & 0.17 \\
SurgVLP & 0.49 & 0.21 & 0.15 & 0.16 & 0.66 & 0.57 & 0.43 & 0.48 & 0.80 \\
Gemini-1.5-Pro & 0.86 & 0.24 & 0.16 & \textbf{0.26} & 0.28 & 0.58 & 0.45 & 0.59 & 0.60 \\
GPT-4o & \textbf{0.89} & \textbf{0.25} & \textbf{0.18} & \textbf{0.26} & 0.28 & \textbf{0.59} & 0.46 & \textbf{0.64} & 0.56 \\
LLaVA-NeXT & 0.64 & 0.18 & 0.14 & 0.14 & 0.47 & 0.58 & 0.46 & 0.47 & 0.86 \\
PaliGemma & 0.76 & 0.02 & 0.01 & 0.08 & 0.11 & 0.02 & 0.01 & 0.18 & 0.02 \\
\end{tabular}
\end{table*}

\begin{table*}
\centering
\caption{Inst C45}\label{tab:InstC45}\setlength{\tabcolsep}{4pt}
\begin{tabular}{lccccc|cccc}
Model & A & F1 & J & P & R & wF1 & wJ & wP & wR \\
\hline
CLIP & 0.35 & 0.28 & 0.21 & 0.21 & \textbf{0.86} & 0.69 & 0.57 & 0.57 & \textbf{1.00} \\
Med-Gemini & 0.78 & 0.33 & 0.24 & 0.44 & 0.31 & \textbf{0.75} & \textbf{0.65} & 0.74 & 0.78 \\
OpenCLIP & 0.44 & 0.28 & 0.22 & 0.22 & 0.79 & 0.71 & 0.59 & 0.60 & 0.97 \\
Qwen2-VL & 0.73 & 0.25 & 0.18 & 0.29 & 0.37 & 0.51 & 0.41 & 0.65 & 0.55 \\
SurgVLP & 0.51 & 0.28 & 0.21 & 0.23 & 0.75 & 0.69 & 0.56 & 0.60 & 0.89 \\
Gemini-1.5-Pro & 0.81 & 0.32 & 0.23 & 0.40 & 0.37 & 0.58 & 0.46 & 0.76 & 0.57 \\
GPT-4o & \textbf{0.87} & \textbf{0.40} & \textbf{0.29} & \textbf{0.45} & 0.38 & 0.70 & 0.57 & \textbf{0.80} & 0.63 \\
LLaVA-NeXT & 0.43 & 0.29 & 0.23 & 0.23 & 0.74 & 0.69 & 0.57 & 0.57 & 0.92 \\
PaliGemma & 0.62 & 0.03 & 0.02 & 0.13 & 0.18 & 0.01 & 0.00 & 0.22 & 0.04 \\
\end{tabular}
\end{table*}

\begin{table*}
\centering
\caption{Suture Ski JS}\label{tab:jigsaws_skill_assessment_debug_Suturing}\setlength{\tabcolsep}{4pt}
\begin{tabular}{lccccc|cccc}
Model & A & F1 & J & P & R & wF1 & wJ & wP & wR \\
\hline
InternVL2 & 0.13 & 0.06 & 0.03 & 0.04 & 0.20 & 0.05 & 0.03 & 0.04 & 0.13 \\
Phi-3.5 Vision & 0.26 & 0.10 & 0.06 & 0.08 & 0.20 & 0.13 & 0.09 & 0.12 & 0.26 \\
Qwen2-VL & \textbf{0.33} & \textbf{0.12} & \textbf{0.08} & 0.09 & 0.22 & \textbf{0.18} & \textbf{0.12} & 0.13 & \textbf{0.33} \\
Gemini-1.5-Pro & 0.18 & 0.10 & 0.06 & \textbf{0.20} & 0.22 & 0.14 & 0.08 & \textbf{0.31} & 0.18 \\
GPT-4o & 0.30 & 0.11 & 0.07 & 0.07 & \textbf{0.22} & 0.16 & 0.11 & 0.11 & 0.30 \\
\end{tabular}
\end{table*}

\begin{table*}
\centering
\caption{Needle Ges JS}\label{tab:NeedleGesJS}\setlength{\tabcolsep}{4pt}
\begin{tabular}{lccccc|cccc}
Model & A & F1 & J & P & R & wF1 & wJ & wP & wR \\
\hline
InternVL2 & 0.06 & 0.01 & 0.01 & 0.01 & 0.10 & 0.01 & 0.00 & 0.00 & 0.06 \\
Phi-3.5 Vision & 0.05 & 0.02 & 0.01 & 0.05 & 0.06 & 0.03 & 0.01 & 0.12 & 0.05 \\
Qwen2-VL & 0.06 & 0.01 & 0.01 & 0.01 & 0.10 & 0.01 & 0.00 & 0.00 & 0.06 \\
Gemini-1.5-Pro & 0.16 & 0.09 & 0.05 & 0.10 & 0.10 & \textbf{0.16} & \textbf{0.09} & \textbf{0.18} & 0.16 \\
GPT-4o & \textbf{0.22} & \textbf{0.10} & \textbf{0.06} & \textbf{0.14} & \textbf{0.13} & 0.15 & \textbf{0.09} & 0.17 & \textbf{0.22} \\
\end{tabular}
\end{table*}

\begin{table*}
\centering
\caption{Anat Rec DS}\label{tab:AnatRecDS}\setlength{\tabcolsep}{4pt}
\begin{tabular}{lccccc|cccc}
Model & A & F1 & J & P & R & wF1 & wJ & wP & wR \\
\hline
CLIP & 0.22 & 0.31 & 0.20 & 0.20 & \textbf{0.98} & 0.43 & 0.29 & 0.29 & \textbf{0.98} \\
Med-Gemini & 0.76 & \textbf{0.33} & \textbf{0.21} & 0.36 & 0.36 & \textbf{0.45} & \textbf{0.30} & \textbf{0.56} & 0.41 \\
OpenCLIP & 0.47 & 0.30 & 0.20 & 0.22 & 0.67 & 0.43 & 0.29 & 0.33 & 0.74 \\
Qwen2-VL & 0.70 & 0.21 & 0.13 & 0.26 & 0.31 & 0.33 & 0.22 & 0.32 & 0.46 \\
SurgVLP & 0.29 & 0.32 & 0.20 & 0.21 & 0.92 & 0.43 & 0.29 & 0.30 & 0.90 \\
Gemini-1.5-Pro & 0.76 & 0.29 & 0.18 & 0.34 & 0.32 & 0.32 & 0.21 & 0.39 & 0.36 \\
GPT-4o & \textbf{0.80} & 0.17 & 0.10 & \textbf{0.43} & 0.13 & 0.20 & 0.12 & 0.53 & 0.16 \\
LLaVA-NeXT & 0.33 & 0.30 & 0.18 & 0.20 & 0.80 & 0.39 & 0.25 & 0.28 & 0.78 \\
PaliGemma & 0.53 & 0.30 & 0.19 & 0.21 & 0.59 & 0.41 & 0.27 & 0.30 & 0.69 \\
\end{tabular}
\end{table*}

\begin{table*}
\centering
\caption{Skill IM}\label{tab:SkillIM}\setlength{\tabcolsep}{4pt}
\begin{tabular}{lccccc|cccc}
Model & A & F1 & J & P & R & wF1 & wJ & wP & wR \\
\hline
InternVL2 & 0.40 & \textbf{0.20} & \textbf{0.13} & 0.19 & \textbf{0.27} & \textbf{0.29} & \textbf{0.20} & 0.27 & 0.40 \\
Phi-3.5 Vision & \textbf{0.41} & 0.19 & 0.13 & 0.17 & 0.27 & 0.29 & 0.20 & 0.24 & \textbf{0.41} \\
Qwen2-VL & 0.40 & 0.18 & 0.12 & \textbf{0.26} & 0.26 & 0.28 & 0.19 & \textbf{0.35} & 0.40 \\
Gemini-1.5-Pro & 0.37 & 0.18 & 0.12 & 0.18 & 0.24 & 0.29 & 0.20 & 0.27 & 0.37 \\
GPT-4o & 0.27 & 0.15 & 0.10 & 0.18 & 0.21 & 0.22 & 0.14 & 0.27 & 0.27 \\
\end{tabular}
\end{table*}

\begin{table*}
\centering
\caption{Severity IM}\label{tab:SeverityIM}\setlength{\tabcolsep}{4pt}
\begin{tabular}{lccccc|cccc}
Model & A & F1 & J & P & R & wF1 & wJ & wP & wR \\
\hline
CLIP & 0.10 & 0.05 & 0.03 & 0.04 & 0.17 & 0.04 & 0.02 & 0.03 & 0.10 \\
OpenCLIP & 0.12 & 0.04 & 0.02 & 0.02 & 0.20 & 0.02 & 0.01 & 0.01 & 0.12 \\
Qwen2-VL & 0.38 & 0.11 & 0.08 & 0.08 & 0.20 & 0.21 & 0.15 & 0.15 & 0.38 \\
SurgVLP & 0.26 & \textbf{0.22} & \textbf{0.13} & \textbf{0.39} & \textbf{0.30} & 0.23 & 0.14 & \textbf{0.43} & 0.26 \\
Gemini-1.5-Pro & \textbf{0.40} & 0.16 & 0.10 & 0.28 & 0.22 & 0.24 & 0.16 & 0.27 & \textbf{0.40} \\
GPT-4o & \textbf{0.40} & 0.20 & \textbf{0.13} & 0.36 & 0.25 & \textbf{0.28} & \textbf{0.18} & 0.33 & \textbf{0.40} \\
LLaVA-NeXT & 0.18 & 0.09 & 0.05 & 0.08 & 0.18 & 0.11 & 0.06 & 0.12 & 0.18 \\
PaliGemma & 0.18 & 0.09 & 0.05 & 0.09 & 0.17 & 0.13 & 0.07 & 0.15 & 0.18 \\
\end{tabular}
\end{table*}

\begin{table*}
\centering
\caption{Knot Ges JS}\label{tab:KnotGesJS}\setlength{\tabcolsep}{4pt}
\begin{tabular}{lccccc|cccc}
Model & A & F1 & J & P & R & wF1 & wJ & wP & wR \\
\hline
InternVL2 & \textbf{0.05} & \textbf{0.01} & \textbf{0.01} & \textbf{0.01} & \textbf{0.14} & \textbf{0.01} & \textbf{0.00} & \textbf{0.00} & \textbf{0.05} \\
Phi-3.5 Vision & \textbf{0.05} & \textbf{0.01} & \textbf{0.01} & \textbf{0.01} & 0.09 & \textbf{0.01} & \textbf{0.00} & \textbf{0.00} & \textbf{0.05} \\
Qwen2-VL & \textbf{0.05} & \textbf{0.01} & \textbf{0.01} & \textbf{0.01} & \textbf{0.14} & 0.00 & \textbf{0.00} & \textbf{0.00} & \textbf{0.05} \\
Gemini-1.5-Pro & 0.03 & \textbf{0.01} & 0.00 & 0.00 & 0.04 & \textbf{0.01} & \textbf{0.00} & \textbf{0.00} & 0.03 \\
GPT-4o & 0.00 & 0.00 & 0.00 & 0.00 & 0.00 & 0.00 & \textbf{0.00} & \textbf{0.00} & 0.00 \\
\end{tabular}
\end{table*}

\begin{table*}
\centering
\caption{E-Clf SST}\label{tab:E-ClfSST}\setlength{\tabcolsep}{4pt}
\begin{tabular}{lccccc|cccc}
Model & A & F1 & J & P & R & wF1 & wJ & wP & wR \\
\hline
InternVL2 & 0.57 & 0.18 & 0.14 & 0.15 & 0.24 & 0.44 & 0.35 & 0.36 & 0.57 \\
Phi-3.5 Vision & 0.58 & 0.25 & 0.18 & 0.24 & 0.27 & 0.52 & 0.39 & 0.47 & 0.58 \\
Qwen2-VL & 0.59 & 0.19 & 0.15 & 0.15 & 0.24 & 0.45 & 0.36 & 0.37 & 0.59 \\
Gemini-1.5-Pro & \textbf{0.62} & 0.36 & 0.26 & 0.45 & 0.36 & \textbf{0.60} & \textbf{0.46} & \textbf{0.64} & \textbf{0.62} \\
GPT-4o & 0.61 & \textbf{0.52} & \textbf{0.38} & \textbf{0.62} & \textbf{0.49} & 0.58 & 0.44 & 0.58 & 0.61 \\
\end{tabular}
\end{table*}

\begin{table*}
\centering
\caption{Phase C80}\label{tab:PhaseC80}\setlength{\tabcolsep}{4pt}
\begin{tabular}{lccccc|cccc}
Model & A & F1 & J & P & R & wF1 & wJ & wP & wR \\
\hline
CLIP & 0.06 & 0.02 & 0.01 & 0.03 & 0.13 & 0.01 & 0.00 & 0.01 & 0.06 \\
Med-Gemini & \textbf{0.43} & 0.23 & 0.15 & 0.24 & 0.31 & \textbf{0.52} & \textbf{0.37} & \textbf{0.76} & \textbf{0.43} \\
OpenCLIP & 0.19 & 0.11 & 0.06 & 0.15 & 0.21 & 0.18 & 0.11 & 0.33 & 0.19 \\
Qwen2-VL & 0.26 & 0.18 & 0.11 & 0.29 & 0.22 & 0.23 & 0.14 & 0.38 & 0.26 \\
SurgVLP & 0.41 & \textbf{0.29} & 0.18 & 0.29 & 0.30 & 0.41 & 0.27 & 0.42 & 0.41 \\
Gemini-1.5-Pro & 0.41 & 0.14 & 0.09 & 0.29 & 0.17 & 0.32 & 0.21 & 0.42 & 0.41 \\
GPT-4o & \textbf{0.43} & \textbf{0.29} & \textbf{0.19} & \textbf{0.47} & 0.29 & 0.37 & 0.25 & 0.41 & \textbf{0.43} \\
LLaVA-NeXT & 0.35 & 0.08 & 0.05 & 0.23 & 0.14 & 0.18 & 0.12 & 0.18 & 0.35 \\
PaliGemma & 0.09 & 0.10 & 0.06 & 0.23 & \textbf{0.32} & 0.04 & 0.02 & 0.10 & 0.40 \\
\end{tabular}
\end{table*}

\begin{table*}
\centering
\caption{Action HC}\label{tab:ActionHC}\setlength{\tabcolsep}{4pt}
\begin{tabular}{lccccc|cccc}
Model & A & F1 & J & P & R & wF1 & wJ & wP & wR \\
\hline
CLIP & 0.21 & \textbf{0.23} & \textbf{0.21} & 0.21 & \textbf{1.00} & \textbf{0.88} & \textbf{0.80} & 0.80 & \textbf{1.00} \\
OpenCLIP & 0.24 & \textbf{0.23} & 0.20 & 0.21 & 0.98 & 0.87 & 0.78 & 0.81 & 0.96 \\
Qwen2-VL & 0.52 & 0.22 & 0.19 & \textbf{0.23} & 0.73 & 0.82 & 0.71 & \textbf{0.86} & 0.80 \\
SurgVLP & 0.21 & \textbf{0.23} & \textbf{0.21} & 0.21 & \textbf{1.00} & \textbf{0.88} & \textbf{0.80} & 0.80 & \textbf{1.00} \\
Gemini-1.5-Pro & 0.51 & 0.12 & 0.07 & 0.19 & 0.48 & 0.41 & 0.26 & 0.73 & 0.30 \\
GPT-4o & \textbf{0.65} & 0.19 & 0.14 & \textbf{0.23} & 0.46 & 0.69 & 0.53 & \textbf{0.86} & 0.58 \\
LLaVA-NeXT & 0.59 & 0.00 & 0.00 & 0.14 & 0.25 & 0.00 & 0.00 & 0.53 & 0.01 \\
PaliGemma & 0.27 & \textbf{0.23} & \textbf{0.21} & 0.22 & 0.98 & \textbf{0.88} & 0.79 & 0.85 & 0.93 \\
\end{tabular}
\end{table*}

\begin{table*}
\centering
\caption{CVS IM}\label{tab:CVSIM}\setlength{\tabcolsep}{4pt}
\begin{tabular}{lccccc|cccc}
Model & A & F1 & J & P & R & wF1 & wJ & wP & wR \\
\hline
CLIP & 0.30 & 0.45 & \textbf{0.30} & 0.30 & \textbf{1.00} & \textbf{0.48} & \textbf{0.32} & 0.32 & \textbf{1.00} \\
OpenCLIP & 0.34 & \textbf{0.46} & \textbf{0.30} & 0.30 & 0.96 & \textbf{0.48} & \textbf{0.32} & 0.32 & 0.97 \\
Qwen2-VL & \textbf{0.71} & 0.01 & 0.00 & \textbf{0.93} & 0.00 & 0.01 & 0.00 & \textbf{0.92} & 0.00 \\
SurgVLP & 0.30 & 0.45 & \textbf{0.30} & 0.30 & \textbf{1.00} & \textbf{0.48} & \textbf{0.32} & 0.32 & 0.99 \\
Gemini-1.5-Pro & \textbf{0.71} & 0.09 & 0.06 & 0.86 & 0.06 & 0.11 & 0.06 & 0.84 & 0.07 \\
GPT-4o & \textbf{0.71} & 0.05 & 0.03 & 0.35 & 0.03 & 0.05 & 0.03 & 0.35 & 0.03 \\
LLaVA-NeXT & 0.70 & 0.00 & 0.00 & 0.00 & 0.00 & 0.00 & 0.00 & 0.00 & 0.00 \\
PaliGemma & 0.51 & 0.35 & 0.23 & 0.23 & 0.67 & 0.41 & 0.28 & 0.28 & 0.79 \\
\end{tabular}
\end{table*}

\begin{table*}
\centering
\caption{Target C45}\label{tab:TargetC45}\setlength{\tabcolsep}{4pt}
\begin{tabular}{lccccc|cccc}
Model & A & F1 & J & P & R & wF1 & wJ & wP & wR \\
\hline
CLIP & 0.09 & 0.14 & 0.09 & 0.09 & \textbf{0.93} & 0.47 & 0.37 & 0.37 & \textbf{1.00} \\
Med-Gemini & 0.67 & 0.14 & \textbf{0.12} & 0.15 & 0.25 & \textbf{0.73} & \textbf{0.62} & \textbf{0.81} & 0.67 \\
OpenCLIP & 0.52 & 0.12 & 0.08 & 0.17 & 0.57 & 0.44 & 0.33 & 0.44 & 0.74 \\
Qwen2-VL & \textbf{0.89} & 0.08 & 0.06 & 0.07 & 0.11 & 0.41 & 0.33 & 0.35 & 0.50 \\
SurgVLP & 0.60 & \textbf{0.16} & 0.10 & 0.12 & 0.54 & 0.46 & 0.34 & 0.39 & 0.71 \\
Gemini-1.5-Pro & 0.88 & 0.14 & 0.09 & \textbf{0.26} & 0.15 & 0.42 & 0.29 & 0.55 & 0.40 \\
GPT-4o & \textbf{0.89} & 0.15 & 0.10 & 0.23 & 0.14 & 0.42 & 0.29 & 0.53 & 0.37 \\
LLaVA-NeXT & 0.72 & 0.08 & 0.06 & 0.06 & 0.29 & 0.41 & 0.33 & 0.33 & 0.66 \\
PaliGemma & 0.85 & 0.02 & 0.01 & 0.13 & 0.07 & 0.06 & 0.03 & 0.46 & 0.03 \\
\end{tabular}
\end{table*}

\begin{table*}
\centering
\caption{CVS ES}\label{tab:CVSES}\setlength{\tabcolsep}{4pt}
\begin{tabular}{lccccc|cccc}
Model & A & F1 & J & P & R & wF1 & wJ & wP & wR \\
\hline
CLIP & 0.23 & 0.37 & 0.23 & 0.23 & \textbf{1.00} & 0.38 & 0.24 & 0.24 & \textbf{1.00} \\
OpenCLIP & 0.47 & \textbf{0.41} & \textbf{0.26} & 0.27 & 0.83 & \textbf{0.42} & \textbf{0.27} & 0.28 & 0.84 \\
Qwen2-VL & \textbf{0.77} & 0.02 & 0.01 & \textbf{0.57} & 0.01 & 0.02 & 0.01 & \textbf{0.58} & 0.01 \\
SurgVLP & 0.26 & 0.38 & 0.23 & 0.23 & 0.99 & 0.39 & 0.24 & 0.24 & 0.99 \\
Gemini-1.5-Pro & 0.76 & 0.08 & 0.05 & 0.26 & 0.06 & 0.08 & 0.05 & 0.26 & 0.06 \\
GPT-4o & 0.74 & 0.10 & 0.05 & 0.23 & 0.07 & 0.10 & 0.05 & 0.23 & 0.07 \\
LLaVA-NeXT & \textbf{0.77} & 0.00 & 0.00 & 0.00 & 0.00 & 0.00 & 0.00 & 0.00 & 0.00 \\
PaliGemma & 0.50 & 0.29 & 0.18 & 0.19 & 0.65 & 0.33 & 0.21 & 0.21 & 0.74 \\
\end{tabular}
\end{table*}

\begin{table*}
\centering
\caption{E-Clf HC}\label{tab:E-ClfHC}\setlength{\tabcolsep}{4pt}
\begin{tabular}{lccccc|cccc}
Model & A & F1 & J & P & R & wF1 & wJ & wP & wR \\
\hline
InternVL2 & 0.38 & \textbf{0.27} & 0.17 & 0.36 & \textbf{0.40} & 0.37 & 0.23 & 0.60 & 0.38 \\
Phi-3.5 Vision & 0.16 & 0.08 & 0.05 & 0.05 & 0.20 & 0.06 & 0.04 & 0.04 & 0.16 \\
Qwen2-VL & 0.24 & 0.16 & 0.09 & \textbf{0.40} & 0.35 & 0.15 & 0.08 & \textbf{0.70} & 0.24 \\
Gemini-1.5-Pro & 0.16 & 0.15 & 0.09 & 0.22 & 0.17 & 0.17 & 0.10 & 0.41 & 0.16 \\
GPT-4o & \textbf{0.44} & 0.26 & \textbf{0.18} & 0.30 & 0.32 & \textbf{0.46} & \textbf{0.31} & 0.64 & \textbf{0.44} \\
\end{tabular}
\end{table*}

\begin{table*}
\centering
\caption{E-Clf C80}\label{tab:E-ClfC80}\setlength{\tabcolsep}{4pt}
\begin{tabular}{lccccc|cccc}
Model & A & F1 & J & P & R & wF1 & wJ & wP & wR \\
\hline
InternVL2 & 0.78 & 0.22 & 0.20 & 0.20 & 0.25 & 0.70 & 0.62 & 0.63 & 0.78 \\
Phi-3.5 Vision & 0.74 & 0.22 & 0.20 & 0.22 & 0.23 & 0.71 & 0.65 & 0.69 & 0.74 \\
Qwen2-VL & \textbf{0.80} & 0.22 & 0.20 & 0.20 & 0.25 & 0.71 & 0.64 & 0.64 & \textbf{0.80} \\
Gemini-1.5-Pro & 0.64 & 0.27 & 0.21 & 0.31 & 0.46 & 0.69 & 0.59 & 0.79 & 0.64 \\
GPT-4o & 0.75 & \textbf{0.37} & \textbf{0.29} & \textbf{0.40} & \textbf{0.55} & \textbf{0.78} & \textbf{0.68} & \textbf{0.84} & 0.75 \\
\end{tabular}
\end{table*}

\begin{table*}
\centering
\caption{Exposure AL}\label{tab:autolaporo_maneuver_classification}\setlength{\tabcolsep}{4pt}
\begin{tabular}{lccccc|cccc}
Model & A & F1 & J & P & R & wF1 & wJ & wP & wR \\
\hline
InternVL2 & 0.07 & 0.02 & 0.01 & 0.01 & 0.14 & 0.01 & 0.00 & 0.00 & 0.07 \\
Phi-3.5 Vision & 0.07 & 0.02 & 0.01 & 0.01 & 0.14 & 0.01 & 0.00 & 0.00 & 0.07 \\
Qwen2-VL & 0.19 & 0.13 & 0.07 & 0.16 & 0.17 & 0.13 & 0.08 & 0.14 & 0.19 \\
Gemini-1.5-Pro & 0.26 & 0.14 & 0.09 & 0.13 & \textbf{0.21} & 0.17 & 0.10 & 0.16 & 0.26 \\
GPT-4o & \textbf{0.29} & \textbf{0.17} & \textbf{0.10} & \textbf{0.22} & 0.21 & \textbf{0.22} & \textbf{0.14} & \textbf{0.28} & \textbf{0.29} \\
\end{tabular}
\end{table*}

\begin{table*}
\centering
\caption{Needle Ski JS}\label{tab:jigsaws_skill_assessment_debug_Needle_Passing}\setlength{\tabcolsep}{4pt}
\begin{tabular}{lccccc|cccc}
Model & A & F1 & J & P & R & wF1 & wJ & wP & wR \\
\hline
InternVL2 & 0.05 & 0.02 & 0.01 & 0.01 & 0.05 & 0.02 & 0.01 & 0.01 & 0.05 \\
Phi-3.5 Vision & \textbf{0.39} & \textbf{0.24} & \textbf{0.16} & \textbf{0.23} & \textbf{0.35} & \textbf{0.32} & \textbf{0.21} & \textbf{0.33} & \textbf{0.39} \\
Qwen2-VL & 0.31 & 0.12 & 0.08 & 0.08 & 0.25 & 0.17 & 0.12 & 0.12 & 0.31 \\
Gemini-1.5-Pro & 0.18 & 0.07 & 0.05 & 0.08 & 0.13 & 0.12 & 0.08 & 0.16 & 0.18 \\
GPT-4o & 0.31 & 0.13 & 0.09 & 0.09 & 0.27 & 0.16 & 0.11 & 0.11 & 0.31 \\
\end{tabular}
\end{table*}

\begin{table*}
\centering
\caption{Tool HC}\label{tab:ToolHC}\setlength{\tabcolsep}{4pt}
\begin{tabular}{lccccc|cccc}
Model & A & F1 & J & P & R & wF1 & wJ & wP & wR \\
\hline
CLIP & 0.58 & 0.22 & 0.16 & 0.22 & 0.47 & 0.60 & 0.46 & 0.53 & 0.76 \\
OpenCLIP & 0.72 & 0.22 & 0.15 & 0.26 & 0.33 & 0.39 & 0.31 & 0.57 & 0.40 \\
Qwen2-VL & 0.72 & 0.32 & 0.24 & 0.32 & 0.56 & \textbf{0.70} & \textbf{0.57} & 0.64 & \textbf{0.84} \\
SurgVLP & 0.41 & 0.25 & 0.18 & 0.20 & \textbf{0.69} & 0.61 & 0.46 & 0.52 & 0.82 \\
Gemini-1.5-Pro & 0.81 & 0.28 & 0.20 & 0.35 & 0.35 & 0.54 & 0.42 & 0.64 & 0.56 \\
GPT-4o & \textbf{0.88} & \textbf{0.36} & \textbf{0.26} & 0.45 & 0.33 & 0.69 & 0.55 & \textbf{0.84} & 0.59 \\
LLaVA-NeXT & 0.50 & 0.26 & 0.19 & 0.24 & 0.59 & 0.63 & 0.49 & 0.64 & 0.69 \\
PaliGemma & 0.11 & 0.35 & \textbf{0.26} & \textbf{0.66} & 0.32 & \textbf{0.70} & 0.56 & 0.83 & 0.65 \\
\end{tabular}
\end{table*}

\begin{table*}
\centering
\caption{Skill HC}\label{tab:SkillHC}\setlength{\tabcolsep}{4pt}
\begin{tabular}{lccccc|cccc}
Model & A & F1 & J & P & R & wF1 & wJ & wP & wR \\
\hline
InternVL2 & \textbf{0.40} & \textbf{0.26} & \textbf{0.17} & \textbf{0.30} & \textbf{0.34} & \textbf{0.30} & \textbf{0.20} & \textbf{0.37} & \textbf{0.40} \\
Phi-3.5 Vision & 0.26 & 0.12 & 0.08 & 0.08 & 0.30 & 0.13 & 0.09 & 0.09 & 0.26 \\
Qwen2-VL & 0.15 & 0.10 & 0.06 & 0.06 & 0.32 & 0.05 & 0.03 & 0.03 & 0.15 \\
Gemini-1.5-Pro & 0.25 & 0.12 & 0.08 & 0.12 & 0.31 & 0.12 & 0.08 & 0.16 & 0.25 \\
GPT-4o & 0.17 & 0.08 & 0.05 & 0.05 & 0.30 & 0.06 & 0.03 & 0.03 & 0.17 \\
\end{tabular}
\end{table*}

\begin{table*}
\centering
\caption{Suture Ges JS}\label{tab:SutureGesJS}\setlength{\tabcolsep}{4pt}
\begin{tabular}{lccccc|cccc}
Model & A & F1 & J & P & R & wF1 & wJ & wP & wR \\
\hline
InternVL2 & 0.04 & 0.01 & 0.00 & 0.09 & 0.09 & 0.01 & 0.00 & 0.21 & 0.04 \\
Phi-3.5 Vision & 0.10 & 0.04 & 0.02 & 0.03 & 0.04 & 0.09 & 0.05 & 0.08 & 0.10 \\
Qwen2-VL & 0.03 & 0.01 & 0.00 & 0.01 & 0.07 & 0.01 & 0.01 & 0.03 & 0.03 \\
Gemini-1.5-Pro & 0.18 & 0.08 & 0.05 & 0.14 & 0.10 & 0.14 & 0.08 & 0.22 & 0.18 \\
GPT-4o & \textbf{0.33} & \textbf{0.10} & \textbf{0.07} & \textbf{0.16} & \textbf{0.14} & \textbf{0.23} & \textbf{0.15} & \textbf{0.31} & \textbf{0.33} \\
\end{tabular}
\end{table*}

\begin{table*}
\centering
\caption{Exposure AL}\label{tab:ExposureAL}\setlength{\tabcolsep}{4pt}
\begin{tabular}{lccccc|cccc}
Model & A & F1 & J & P & R & wF1 & wJ & wP & wR \\
\hline
InternVL2 & 0.07 & 0.02 & 0.01 & 0.01 & 0.14 & 0.01 & 0.00 & 0.00 & 0.07 \\
Phi-3.5 Vision & 0.07 & 0.02 & 0.01 & 0.01 & 0.14 & 0.01 & 0.00 & 0.00 & 0.07 \\
Qwen2-VL & 0.19 & 0.13 & 0.07 & 0.16 & 0.17 & 0.13 & 0.08 & 0.14 & 0.19 \\
Gemini-1.5-Pro & 0.26 & 0.14 & 0.09 & 0.13 & \textbf{0.21} & 0.17 & 0.10 & 0.16 & 0.26 \\
GPT-4o & \textbf{0.29} & \textbf{0.17} & \textbf{0.10} & \textbf{0.22} & 0.21 & \textbf{0.22} & \textbf{0.14} & \textbf{0.28} & \textbf{0.29} \\
\end{tabular}
\end{table*}

\begin{table*}
\centering
\caption{Phase HC}\label{tab:PhaseHC}\setlength{\tabcolsep}{4pt}
\begin{tabular}{lccccc|cccc}
Model & A & F1 & J & P & R & wF1 & wJ & wP & wR \\
\hline
CLIP & 0.14 & 0.05 & 0.03 & 0.11 & 0.14 & 0.04 & 0.02 & 0.04 & 0.14 \\
OpenCLIP & 0.15 & 0.11 & 0.06 & 0.15 & 0.18 & 0.18 & 0.10 & 0.33 & 0.15 \\
Qwen2-VL & 0.17 & 0.12 & 0.07 & 0.28 & 0.20 & 0.12 & 0.07 & 0.45 & 0.17 \\
SurgVLP & 0.26 & 0.24 & 0.14 & 0.28 & 0.29 & 0.27 & 0.15 & 0.48 & 0.26 \\
Gemini-1.5-Pro & 0.44 & 0.21 & 0.13 & 0.31 & 0.21 & 0.39 & 0.28 & 0.43 & 0.44 \\
GPT-4o & \textbf{0.50} & \textbf{0.33} & \textbf{0.22} & \textbf{0.38} & \textbf{0.33} & \textbf{0.49} & \textbf{0.35} & \textbf{0.52} & \textbf{0.50} \\
LLaVA-NeXT & 0.16 & 0.04 & 0.02 & 0.16 & 0.14 & 0.05 & 0.03 & 0.31 & 0.16 \\
PaliGemma & 0.11 & 0.10 & 0.05 & 0.25 & 0.26 & 0.04 & 0.02 & 0.11 & 0.20 \\
\end{tabular}
\end{table*}

\begin{table*}
\centering
\caption{Action AV}\label{tab:ActionAV}\setlength{\tabcolsep}{4pt}
\begin{tabular}{lccccc|cccc}
Model & A & F1 & J & P & R & wF1 & wJ & wP & wR \\
\hline
CLIP & 0.16 & 0.11 & 0.06 & 0.16 & 0.19 & 0.10 & 0.06 & 0.26 & 0.16 \\
Med-Gemini & \textbf{0.57} & \textbf{0.39} & \textbf{0.26} & \textbf{0.40} & \textbf{0.42} & \textbf{0.58} & \textbf{0.44} & \textbf{0.63} & \textbf{0.57} \\
OpenCLIP & 0.10 & 0.05 & 0.03 & 0.05 & 0.24 & 0.03 & 0.01 & 0.04 & 0.10 \\
Qwen2-VL & 0.49 & 0.21 & 0.15 & 0.23 & 0.25 & 0.41 & 0.31 & 0.41 & 0.49 \\
SurgVLP & 0.37 & 0.25 & 0.16 & 0.26 & 0.26 & 0.39 & 0.26 & 0.43 & 0.37 \\
Gemini-1.5-Pro & 0.41 & 0.29 & 0.18 & 0.29 & 0.30 & 0.43 & 0.29 & 0.48 & 0.41 \\
GPT-4o & 0.48 & 0.33 & 0.22 & \textbf{0.40} & 0.37 & 0.48 & 0.33 & 0.60 & 0.48 \\
LLaVA-NeXT & \textbf{0.57} & 0.18 & 0.14 & 0.14 & 0.25 & 0.41 & 0.32 & 0.32 & \textbf{0.57} \\
PaliGemma & 0.23 & 0.14 & 0.09 & 0.22 & 0.28 & 0.24 & 0.14 & 0.44 & 0.23 \\
\end{tabular}
\end{table*}

\begin{table*}
\centering
\caption{E-Det SST}\label{tab:E-DetSST}\setlength{\tabcolsep}{20pt}
\begin{tabular}{lc}
Model & mIoU \\
\hline
InternVL2 & 0.02 \\
Qwen2-VL & \textbf{0.13} \\
Gemini-1.5-Pro & 0.04 \\
GPT-4o & 0.07 \\
\end{tabular}
\end{table*}

\begin{table*}
\centering
\caption{E-Det C80}\label{tab:E-DetC80}\setlength{\tabcolsep}{20pt}
\begin{tabular}{lc}
Model & mIoU \\
\hline
InternVL2 & 0.09 \\
Qwen2-VL & 0.11 \\
Gemini-1.5-Pro & \textbf{0.15} \\
GPT-4o & 0.12 \\
\end{tabular}
\end{table*}

\begin{table*}
\centering
\caption{Tool SAR}\label{tab:ToolSAR}\setlength{\tabcolsep}{20pt}
\begin{tabular}{lc}
Model & mIoU \\
\hline
MedSAM & 0.54 \\
SAM2 & \textbf{0.60} \\
PaliGemma & 0.13 \\
\end{tabular}
\end{table*}

\begin{table*}
\centering
\caption{Hand AV}\label{tab:HandAV}\setlength{\tabcolsep}{20pt}
\begin{tabular}{lcc}
Model & AP@0.50:0.95 & wAP@0.50:0.95 \\
\hline
Gemini-1.5-Pro & \textbf{0.29} & \textbf{0.29} \\
PaliGemma & 0.08 & 0.08 \\
\end{tabular}
\end{table*}

\begin{table*}
\centering
\caption{Anat Seg DS}\label{tab:AnatSegDS}\setlength{\tabcolsep}{20pt}
\begin{tabular}{lc}
Model & mIoU \\
\hline
MedSAM & 0.70 \\
SAM2 & \textbf{0.71} \\
PaliGemma & 0.13 \\
\end{tabular}
\end{table*}

\begin{table*}
\centering
\caption{Anat ES}\label{tab:AnatES}\setlength{\tabcolsep}{20pt}
\begin{tabular}{lcc}
Model & AP@0.50:0.95 & wAP@0.50:0.95 \\
\hline
Gemini-1.5-Pro & \textbf{0.02} & \textbf{0.03} \\
PaliGemma & 0.00 & 0.00 \\
\end{tabular}
\end{table*}

\begin{table*}
\centering
\caption{Tool ES}\label{tab:ToolES}\setlength{\tabcolsep}{20pt}
\begin{tabular}{lcc}
Model & AP@0.50:0.95 & wAP@0.50:0.95 \\
\hline
Gemini-1.5-Pro & \textbf{0.43} & \textbf{0.43} \\
PaliGemma & 0.23 & 0.23 \\
\end{tabular}
\end{table*}

\begin{table*}
\centering
\caption{Anat AL}\label{tab:AnatAL}\setlength{\tabcolsep}{20pt}
\begin{tabular}{lc}
Model & mIoU \\
\hline
MedSAM & 0.83 \\
SAM2 & \textbf{0.86} \\
PaliGemma & 0.09 \\
\end{tabular}
\end{table*}

\begin{table*}
\centering
\caption{Tool AV}\label{tab:ToolAV}\setlength{\tabcolsep}{20pt}
\begin{tabular}{lcc}
Model & AP@0.50:0.95 & wAP@0.50:0.95 \\
\hline
Gemini-1.5-Pro & \textbf{0.03} & \textbf{0.02} \\
PaliGemma & 0.00 & 0.00 \\
\end{tabular}
\end{table*}

\begin{table*}
\centering
\caption{E-Det HC}\label{tab:E-DetHC}\setlength{\tabcolsep}{20pt}
\begin{tabular}{lc}
Model & mIoU \\
\hline
InternVL2 & 0.07 \\
Qwen2-VL & 0.11 \\
Gemini-1.5-Pro & \textbf{0.25} \\
GPT-4o & 0.12 \\
\end{tabular}
\end{table*}

\begin{table*}
\centering
\caption{E-Det IM}\label{tab:E-DetIM}\setlength{\tabcolsep}{20pt}
\begin{tabular}{lc}
Model & mIoU \\
\hline
InternVL2 & 0.04 \\
Qwen2-VL & \textbf{0.17} \\
Gemini-1.5-Pro & 0.08 \\
GPT-4o & 0.13 \\
\end{tabular}
\end{table*}

\end{document}